\documentclass[10pt,twocolumn,letterpaper]{article}

\PassOptionsToPackage{table,xcdraw}{xcolor}
\usepackage[pagenumbers]{iccv} 
\definecolor{customblue}{HTML}{156082}

\usepackage[normalem]{ulem}
\useunder{\uline}{\ul}{}
\usepackage{diagbox}
\usepackage{multirow}
\usepackage{pifont}
\usepackage[accsupp]{axessibility}
\usepackage[toc,page]{appendix}

\newcommand\minisection[1]{\vspace{2mm}\noindent \textbf{#1}}

\definecolor{Crimson}{HTML}{DC143C}

\definecolor{iccvblue}{rgb}{0.21,0.49,0.74}
\usepackage[pagebackref,breaklinks,colorlinks,allcolors=iccvblue]{hyperref}

\def\confName{ICCV}
\def\confYear{2025}

\title{Quantifying and Narrowing the Unknown: Interactive Text-to-Video Retrieval \\ via Uncertainty Minimization}

\author{Bingqing Zhang$^{1, 2}$\quad Zhuo Cao$^1$\quad Heming Du$^1$\quad Yang Li$^2$\quad Xue Li$^1$\thanks{Corresponding authors}\quad Jiajun Liu$^{2, 1*}$\quad Sen Wang$^{1*}$ 
\\
$^1$ {The University of Queensland, Australia} \\
$^2$ {CSIRO Data61, Australia} \\
{\tt\small  \{bingqing.zhang, william.cao, heming.du\}@uq.edu.au, yang.li1@csiro.au} \\
{\tt\small xueli@eesc.uq.edu.au, jiajun.liu@csiro.au, sen.wang@uq.edu.au}
}

\begin{document}
\maketitle
\begin{abstract}
Despite recent advances, Text-to-video retrieval (TVR) is still hindered by multiple inherent uncertainties, such as ambiguous textual queries, indistinct text-video mappings, and low-quality video frames. Although interactive systems have emerged to address these challenges by refining user intent through clarifying questions, current methods typically rely on heuristic or ad-hoc strategies without explicitly quantifying these uncertainties, limiting their effectiveness. Motivated by this gap, we propose UMIVR, an Uncertainty-Minimizing Interactive Text-to-Video Retrieval framework that explicitly quantifies three critical uncertainties-text ambiguity, mapping uncertainty, and frame uncertainty-via principled, training-free metrics: semantic entropy-based Text Ambiguity Score (TAS), Jensen-Shannon divergence-based Mapping Uncertainty Score (MUS), and a Temporal Quality-based Frame Sampler (TQFS). By adaptively generating targeted clarifying questions guided by these uncertainty measures, UMIVR iteratively refines user queries, significantly reducing retrieval ambiguity. Extensive experiments on multiple benchmarks validate UMIVR's effectiveness, achieving notable gains in Recall@1 (69.2\% after 10 interactive rounds) on the MSR-VTT-1k dataset, thereby establishing an uncertainty-minimizing foundation for interactive TVR. Code will be avaliable at \url{https://github.com/bingqingzhang/umivr}.
\end{abstract}    
\begin{figure}[t]
    \centering
    \includegraphics[width=0.49\textwidth]{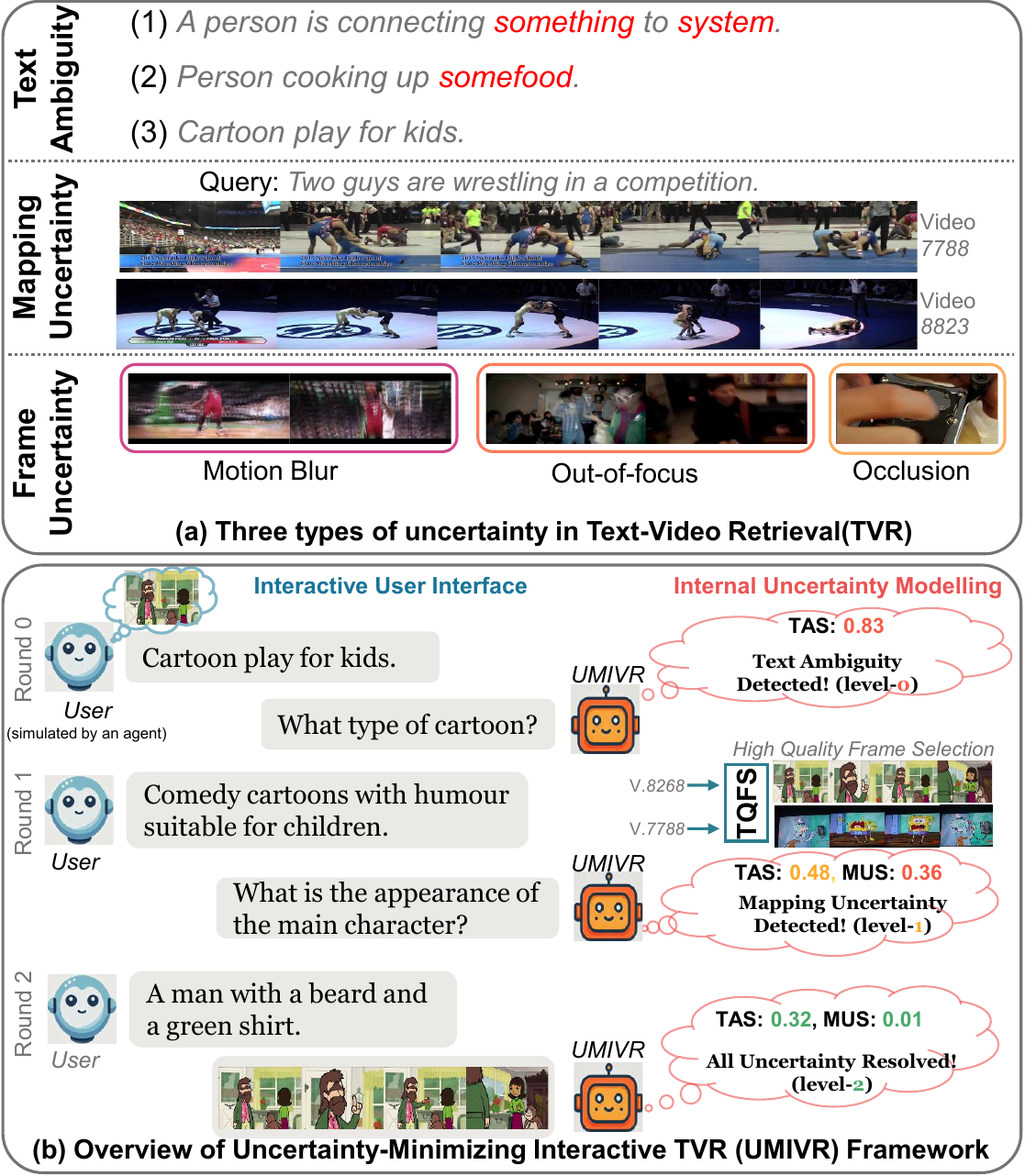}
    \caption{Illustration of uncertainty challenges in Text-to-Video Retrieval (TVR) and our proposed UMIVR framework. \textbf{(a)} Three types of uncertainty that commonly degrade retrieval performance. \textbf{(b)} UMIVR explicitly quantifies all three types of uncertainties per interaction round: Text Ambiguity Score (TAS), Mapping Uncertainty Score (MUS), and Frame Uncertainty (addressed via TQFS for selecting high-quality frames). UMIVR then iteratively generates adaptive clarifying questions, progressively reducing these uncertainties to achieve precise video retrieval.}
    \label{fig:intro}
\end{figure}

\section{Introduction}
\label{sec:intro}

Text-to-Video Retrieval (TVR) has emerged as a crucial task that bridges computer vision and natural language processing, aiming to retrieve relevant video content based on textual queries. Owing to its broad applicability in video search and recommendation, TVR has rapidly evolved from early attention-based mechanisms \cite{CE, MMT} to vision–language pretraining models \cite{clip, blip}. This research paradigm is now diverse, encompassing streamlined encoder designs \cite{clip4clip, clipvip}, advanced training strategies \cite{Fang2023MaskTR}, and improved feature alignment between text and video \cite{xclip, ucofia, Zhang_2025_WACV}. These innovations demonstrate significant performance gains in various evaluations.

Despite notable advances, TVR still remains challenging due to various forms of uncertainty that arise from both textual and visual sources, as illustrated in Fig.~\ref{fig:intro}(a). First, \textbf{text ambiguity} poses a persistent obstacle: textual queries can often be vague, incomplete, or contain polysemous words (e.g.,``something," ``somefood," or very generic phrases like ``Cartoon play for kids"), leading to underspecified retrieval targets. Second, \textbf{mapping uncertainty} highlights that even clearly formulated textual queries (e.g., ``Two guys are wrestling in a competition") can correspond to multiple plausible videos within a large and diverse dataset, making it difficult to pinpoint the most relevant candidate among visually similar alternatives. Third, \textbf{frame uncertainty} arises from deteriorated video frames—such as motion blur, out-of-focus shots, or occlusion of key objects—which obscure important visual cues essential for accurate retrieval. Taken together, these intertwined uncertainties largely degrade retrieval performance, underscoring the necessity for robust mechanisms that explicitly handle ambiguity and noise in both textual and visual domains.

In fact, uncertainty is by no means exclusive to TVR; it is a pervasive challenge in machine learning, encompassing both epistemic and aleatoric uncertainties \cite{uncertainty_survey}. To tackle such uncertainties, user-centric approaches such as active learning methods \cite{acti_learn}, interactive dialogue systems \cite{RahmaniWANY24}, and minimal human supervision \cite{Gupta2024ClusteringIF} have shown remarkable effectiveness. Building on these insights, recent interactive TVR methods \cite{ivr_base,vired,merlin} also recognize the significance of uncertainty. These systems typically employ VideoQA models \cite{vid2sum,blip} or large language models (LLMs) \cite{gpt4o,t0,bart} to generate follow-up questions and simulate user responses, refining queries based on user feedback. However, a key limitation is that existing interactive methods do not explicitly model or quantify uncertainty, relying instead on heuristic or ad-hoc question generation strategies that may not optimally address uncertainties at play.

Motivated by this limitation, we propose a principled approach that directly addresses the uncertainty challenge in TVR. Rather than resorting to ad-hoc techniques, our method systematically quantifies three critical types of uncertainty—text ambiguity, mapping uncertainty, and frame uncertainty—by leveraging semantic entropy, JS divergence, and a novel Temporal Quality-based Frame Sampler (TQFS), respectively, all with \textit{training-free} approaches. Building upon these quantified uncertainties, we further introduce the Uncertainty-Minimizing Interactive Text-to-Video Retrieval (UMIVR) framework, illustrated in Fig.~\ref{fig:intro}(b). Specifically, UMIVR explicitly tracks and updates uncertainty scores (e.g., TAS for text ambiguity and MUS for mapping uncertainty) at each interaction round and adaptively generates clarifying questions tailored to progressively reducing these uncertainties. Through iterative and targeted user interactions, UMIVR effectively mitigates the negative impacts of ambiguous and noisy inputs, systematically refining query precision and ultimately enhancing retrieval accuracy.

We validate the effectiveness of our approach through extensive experiments on several challenging benchmarks. UMIVR consistently outperforms interactive baselines, notably surpassing the non-interactive leading methods by achieving a Hit@1 of 68.9\% on MSR-VTT-1k after only 3 interaction rounds. Furthermore, extensive experiments also demonstrate substantial performance improvements across AVSD, MSVD, and ActivityNet datasets through iterative query clarifying. Beyond its empirical strength, UMIVR exhibits remarkable generalizability, as the proposed TQFS module can readily serve as a plug-in enhancement for existing TVR models, and the UMIVR architecture seamlessly extends to interactive text-to-image retrieval scenarios, underscoring its broad applicability across multimodal retrieval tasks.

In summary, our contributions are three-fold: 
\begin{itemize}[leftmargin=*, nolistsep]
    \item{
    We explicitly identify the uncertainty challenges in TVR and propose quantitative metrics tailored to distinct uncertainty types, thereby providing a more rigorous framework for understanding retrieval ambiguity;
    }
    \item{We introduce the UMIVR framework, which unifies video retrieval, captioning, and question answering into an integrated system that leverages uncertainty metrics to enhance query refinement;}
    \item {Through comprehensive experiments, we demonstrate the effectiveness and generalizability of our approach, setting new benchmarks and opening avenues for further research in interactive multimodal retrieval.}
\end{itemize}

\section{Related Work}
\label{sec:literature}

\subsection{Text-to-Video Retrieval}

Text-to-Video Retrieval (TVR) aims at retrieving relevant video content given textual queries via cross-modal alignment. Early methods employed attention-based aggregation of multimodal features~\cite{CE, MMT}, while subsequent approaches improved representation learning~\cite{support_set, teachtext}. Later, pretraining models~\cite{clip, blip} advanced TVR by adapting pretrained image-text encoders to videos~\cite{clip4clip, centerclip} and refining alignment with fine-grained contrastive learning~\cite{xclip, ts2net} and auxiliary captioning tasks~\cite{clipvip,zhao2025continual,zhao2025synthetic}.

Recent studies also highlight the critical challenge posed by uncertainty in TVR. TAM~\cite{tam} and UATVR~\cite{uatvr} approached mapping uncertainty through adaptive visual prototypes and probabilistic embeddings, respectively; PAU~\cite{pau} focused on modeling text ambiguity via evidential theory. Nevertheless, existing approaches consider only single uncertainty aspects, thus providing limited performance gains. In contrast, our approach explicitly identifies and systematically quantifies three key types of uncertainty (text ambiguity, mapping uncertainty, and frame uncertainty) within a unified, training-free interactive retrieval framework, significantly mitigating ambiguity and enhancing retrieval effectiveness.

\subsection{Interactive Vision Retrieval}

Interactive retrieval historically aimed at bridging the semantic gap by leveraging iterative user feedback. Early content-based image retrieval relied on low-level features and relevance feedback to incrementally refine user queries~\cite{relevance_feedback, willhunter}. Deep learning advances subsequently introduced reinforcement learning or zero-shot learning~\cite{chen2021semantics,chen2025svip} for adaptive question generation and query refinement~\cite{ask_confirm, cooperative_dialog, learning_video_questions}. More recently, LLMs such as ChatGPT~\cite{chatgpt} have greatly enhanced interactive retrieval by generating context-aware clarifying questions without additional training~\cite{chatir, plugir, merlin}, achieving remarkable generalization. Despite their success, existing methods primarily rely on heuristic or context-driven question generation, overlooking explicit modeling and quantification of underlying uncertainties. In contrast, we propose UMIVR, an explicitly uncertainty-aware interactive retrieval framework that systematically quantifies and reduces multiple uncertainty types (text ambiguity, mapping uncertainty, frame uncertainty), improving retrieval robustness and accuracy.

\section{Method}

In this section, we present our proposed UMIVR framework, explicitly designed to address the three types of uncertainty identified in Sec. Specifically, we introduce principled metrics for quantifying \emph{text ambiguity} (Sec.~\ref{sec:tas}) and \emph{mapping uncertainty} (Sec.~\ref{sec:mus}), and propose a Temporal Quality-based Frame Sampler (TQFS, Sec.~\ref{sec:tqfs}) to mitigate \emph{frame-level uncertainty}. Finally, we integrate these components into a unified interactive retrieval pipeline (Sec.~\ref{sec:framework}), which generates adaptive clarifying questions based on quantified uncertainties, enabling iterative refinement of user queries and enhancing retrieval precision.

\subsection{Text Ambiguity Score via Semantic Entropy}
\label{sec:tas}

Text ambiguity arises when queries are vague, incomplete, or permit multiple semantic interpretations. Unlike conventional token-level heuristics that often overestimate ambiguity by treating lexical variants separately, semantic entropy~\cite{seman_entropy_iclr,seman_entropy_nat} more accurately captures genuine semantic variability by analyzing distributions of meanings.

Motivated by this insight, we introduce a \emph{Text Ambiguity Score} (TAS) to quantify the semantic uncertainty associated with textual queries. Specifically, we first apply a captioning model to videos in the retrieval database, resulting in a corpus of textual descriptions $\mathcal{C} = \{s_i\}_{i=1}^N$, each describing video $i$. These captions are encoded into normalized embeddings $\mathbf{e}_{s_i}$, which we store offline.

Given a query $x$, we compute its embedding $\mathbf{e}_x$ and retrieve the top-$K$ most similar captions from $\mathcal{C}$. To reduce redundancy and avoid inflated entropy, we cluster these captions into $M$ coherent groups $\{c_j\}_{j=1}^{M}$. For each cluster $c_j$, the aggregated probability is computed as $p(c_j \mid x) = \sum_{c \in c_j} \text{sim}(x, c) \big/ \sum_{k=1}^{M} \sum_{c \in c_k} \text{sim}(x, c)$, reflecting its share of the total similarity mass. 
The semantic entropy is then defined as:
\begin{equation}
SE(x) = -\sum_{j=1}^{M} p(c_j \mid x)\,\log p(c_j \mid x),
\end{equation}
which is further transformed into the final TAS value in $[0,1]$ through a normalization function. This step can also incorporate additional adjustments, such as accounting for the structural complexity of the query, ensuring that more complex, well-specified queries are assigned lower TAS values. Higher TAS indicates greater semantic uncertainty, reflecting more ambiguous textual queries.

\subsection{Mapping Uncertainty Score via JS Divergence}
\label{sec:mus}

Mapping uncertainty arises when similarity scores between a text query and candidate videos lack a clear peak, leading to ambiguous mappings~\cite{uatvr}. To quantify this uncertainty, we propose a Mapping Uncertainty Score (MUS) based on Jensen–Shannon (JS) divergence~\cite{js-divergence}, measuring the deviation of the similarity distribution from an ideal, perfectly certain scenario.

Given top-$k$ similarity scores between a query text and video candidates $[s_1, s_2, \dots, s_k]$ (sorted in descending order), we first transform them into a normalized probability distribution $p$:
\begin{equation}
p_i = \frac{\max(s_i - \bar{s}, 0)^2}{\sum_{j=1}^{k} \max(s_j - \bar{s}, 0)^2},
\end{equation}
where $\bar{s}$ denotes the mean similarity score. This transformation emphasizes high-confidence candidates while suppressing low-confidence noise. If all scores fall below $\bar{s}$, we default to a uniform distribution.

Next, we define an ideal one-hot distribution $q$, representing complete certainty:
\begin{equation}
q_i = 
\begin{cases}
1, & \text{if } i = 1, \\
0, & \text{otherwise}.
\end{cases}
\end{equation}
We then compute the JS divergence between the distributions $p$ and $q$:
\begin{equation}
JSD(p \parallel q) = \frac{1}{2}\text{KL}(p \parallel m) + \frac{1}{2}\text{KL}(q \parallel m),
\end{equation}
where $m = \frac{1}{2}(p + q)$ and $\text{KL}(\cdot)$ indicates Kullback–Leibler divergence~\cite{kl-divergence}.

Finally, we normalize the JS divergence by its theoretical maximum $JSD_{\max}$ to obtain MUS within $[0,1]$:
\begin{equation}
\text{MUS}(x) = \frac{JSD(p \parallel q)}{JSD_{\max}}.
\end{equation}

A higher MUS indicates greater ambiguity in the mapping between query and video candidates, while a lower MUS reflects a more confident retrieval scenario.

\begin{figure}[tbp]
    \centering
    \includegraphics[width=0.95\linewidth]{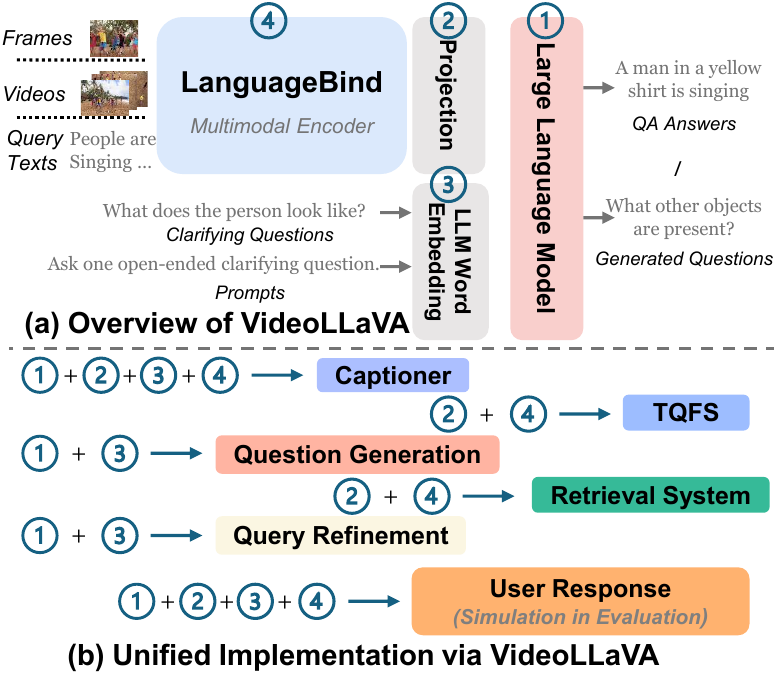}
    \caption{\textbf{A unified implementation of UMIVR with VideoLLaVA.} \textbf{(a)} VideoLLaVA integrates LanguageBind and LLM into a unified architecture, enabling simultaneous handling of video-text tasks. \textbf{(b)} A summary illustrating how UMIVR leverages the single, unified VideoLLaVA model to compactly realize all its core functionalities, significantly simplifying the system architecture compared to prior ensemble or hybrid approaches.}
    \label{fig:component}
\end{figure}

\begin{figure*}[tbp]
    \centering
    \includegraphics[width=1.0\linewidth]{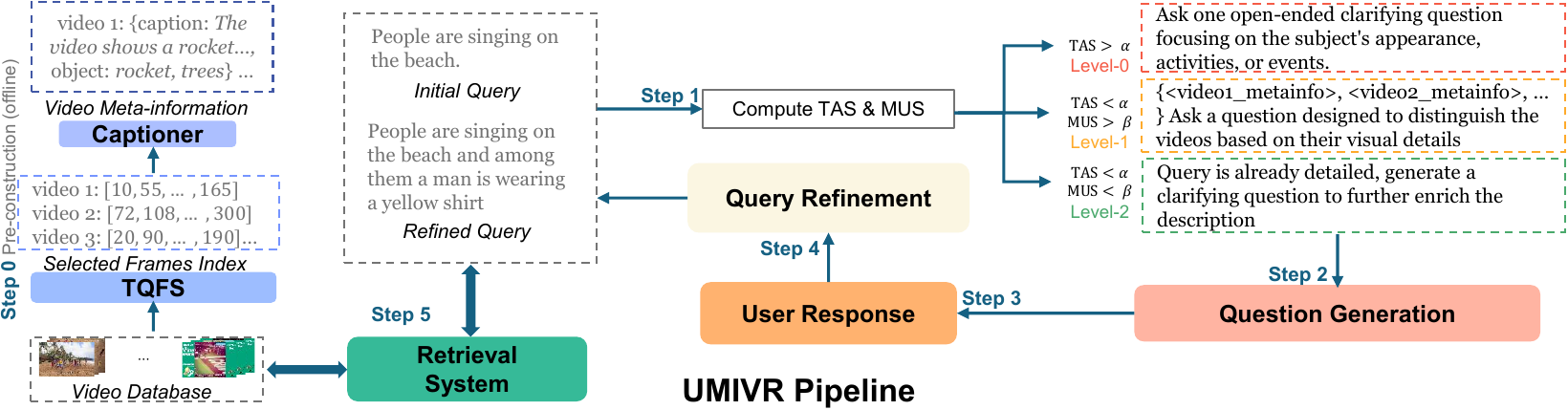}
    \caption{\textbf{Pipeline of the UMIVR framework} Videos are first preprocessed offline (Step 0) by TQFS for high-quality frame selection and captioning to generate meta-information. Given an initial user query, UMIVR quantifies textual and mapping uncertainties (TAS \& MUS, Step 1), adaptively generates clarifying questions at different uncertainty levels (Level-0, 1 and 2, Step 2), and iteratively refines queries based on user responses (Steps 3–4), ultimately retrieving the most relevant videos (Step 5). }
    \label{fig:pipeline}
\end{figure*}

\subsection{Temporal Quality-based Frame Sampler}
\label{sec:tqfs}

Existing video-based methods, such as video retrieval~\cite{clip4clip, clipvip} and video QA models~\cite{blip, videollava}, typically sample frames uniformly, which may inadvertently include low-quality frames affected by defocus or blur~\cite{fgfa, zhang2023object}. To address this issue, we propose a plug-and-play Temporal Quality-based Frame Sampler (TQFS) that adaptively selects high-quality frames while ensuring sufficient temporal coverage.

Given a video $V$ of length $T$ and original frame rate $r$, we first uniformly sample frames at a reduced frame rate $r'$, resulting in $N=\lfloor T\times r/r'\rfloor$ frames $\{F_1, \dots, F_N\}$, each associated with a timestamp $t_i$. Next, we evaluate the visual clarity of each frame $F_i$ using a no-reference image quality assessment (NR-IQA) algorithm $Q(\cdot)$, such as simple Laplacian-variance measures \cite{petrou2010image} or more advanced NR-IQA methods (e.g., BRISQUE\cite{brisque}), assigning each frame a quality score $Q(F_i)$.

To maintain temporal coverage, TQFS divides the video into $M$ uniform temporal bins. In each bin $\mathcal{I}_m$, we select the highest-quality frame:
\begin{equation}
F_m^* = \arg\max_{F_i \in \mathcal{I}_m} Q(F_i),
\end{equation}
resulting in candidate frames \(\{F_1^*, \dots, F_M^*\}\).

To further reduce redundancy and ensure semantic diversity, we extract semantic embeddings $\phi(F_m^*)$ for each candidate frame, forming the embedding matrix $\Phi=[\phi(F_1^*),\dots,\phi(F_M^*)]$. We then apply $K$-means clustering on these embeddings, selecting the highest-quality frame within each cluster. Finally, the selected frames are chronologically ordered, yielding the final $K$ high-quality frames.

Overall, TQFS reduces frame-level uncertainty by emphasizing visually clear and semantically diverse frames, significantly enhancing the robustness of subsequent video retrieval tasks.

\subsection{UMIVR: Uncertainty-Minimizing Interactive Text-to-Video Retrieval Framework}  
\label{sec:framework}

Fig.~\ref{fig:component} and Fig.~\ref{fig:pipeline}
illustrate the detailed architecture and workflow of our proposed UMIVR framework. UMIVR seamlessly integrates text-video retrieval, video captioning, and video question answering into a unified multimodal system, with two key innovations: (1) the adoption of a unified Video-LLM architecture for efficient multi-task integration, and (2) a principled uncertainty-minimizing interactive retrieval pipeline that adaptively generates clarifying questions based on explicitly quantified uncertainties.

\minisection{Unified Video-LLM Architecture.}  
Existing interactive TVR methods typically rely on either multi-model ensemble architectures (e.g., combining BLIP \cite{blip} for retrieval with T0++ \cite{t0} for dialogue generation) or hybrid local/cloud schemes (e.g., PlugIR \cite{plugir} via ChatGPT \cite{chatgpt}), both of which lead to substantial memory overhead or inference latency. To overcome these limitations, UMIVR leverages VideoLLaVA \cite{videollava}, a unified multimodal model that integrates LanguageBind \cite{languagebind}—a robust multimodal encoder—and a LLM via an efficient \textit{align-before-projection} design, as depicted in Fig.~\ref{fig:component}(a). Specifically, visual inputs (frames and videos) and textual queries are first encoded by LanguageBind, projecting visual modality information into the shared language embedding space for accurate cross-modal alignment. In contrast, textual inputs intended for LLM generation tasks, such as clarifying questions and prompts, are directly encoded by the LLM's word embedding layer, enabling effective language generation and comprehension within the model. Fig.~\ref{fig:component}(b) summarizes this unified implementation, highlighting how UMIVR compactly realizes captioning, response simulation, clarifying question generation, and retrieval functionalities within a single model. This unified design not only eliminates cross-model compatibility issues but also significantly reduces memory usage compared to traditional ensemble-based approaches.

\minisection{UMIVR Framework Pipeline.}  
The overall pipeline of UMIVR is depicted in Fig.~\ref{fig:pipeline}. Initially, UMIVR preprocesses the video database offline (\textcolor{customblue}{Step 0}), applying the Temporal Quality-based Frame Sampler (TQFS) to select temporally representative and visually high-quality frames. These selected frames are then passed through VideoLLaVA to generate textual meta-information—including video descriptions and salient object annotations—that will be stored and used during online interactions.

Upon receiving an initial textual query from a user, UMIVR first quantifies the query's uncertainty by computing two complementary uncertainty metrics (\textcolor{customblue}{Step 1}): TAS (Sec.~\ref{sec:tas}) and MUS (Sec.~\ref{sec:mus}). According to these scores, UMIVR adaptively generates clarifying questions at three uncertainty levels (\textcolor{customblue}{Step 2}), effectively guiding user interactions toward reducing ambiguity. Specifically, if TAS is high (greater than a predefined threshold $\alpha$), indicating significant semantic ambiguity in the textual query, the framework instructs VideoLLaVA to generate an open-ended clarifying question, prompting the user for additional context about appearance, activities, or events. Conversely, if TAS is low but MUS remains high (greater than threshold $\beta$), indicating clearly expressed but visually indistinguishable queries, UMIVR leverages the retrieved candidate videos' meta-information (captions, objects, etc.) to generate targeted clarifying questions that explicitly distinguish visually similar videos. Finally, if both uncertainty measures are below their respective thresholds (low uncertainty), UMIVR generates enrichment-oriented questions merely to further enrich the query's descriptive power.

After generating the clarifying question in Step 2, UMIVR expects a user response to refine the query. This interaction can be conducted with real users or approximated for evaluation purposes using simulated responses derived from VideoQA modules (\textcolor{customblue}{Step 3}).
The collected answer is then combined with the initial query via a standard query refinement strategy, yielding a more precise query (\textcolor{customblue}{Step 4}). This refined query significantly reduces uncertainties, enabling a focused and accurate video retrieval (\textcolor{customblue}{Step 5}).

This entire interactive retrieval process is inherently iterative and uncertainty-driven. Each subsequent interaction round benefits from progressively reduced uncertainty scores, systematically refining the query until the retrieved videos closely align with the user's refined intent. By explicitly quantifying uncertainty and adaptively generating clarifying questions, UMIVR robustly addresses ambiguity and noise inherent in text-to-video retrieval, significantly enhancing retrieval accuracy and user satisfaction.

\begin{figure*}[tbp]
    \centering
    \includegraphics[width=\linewidth]{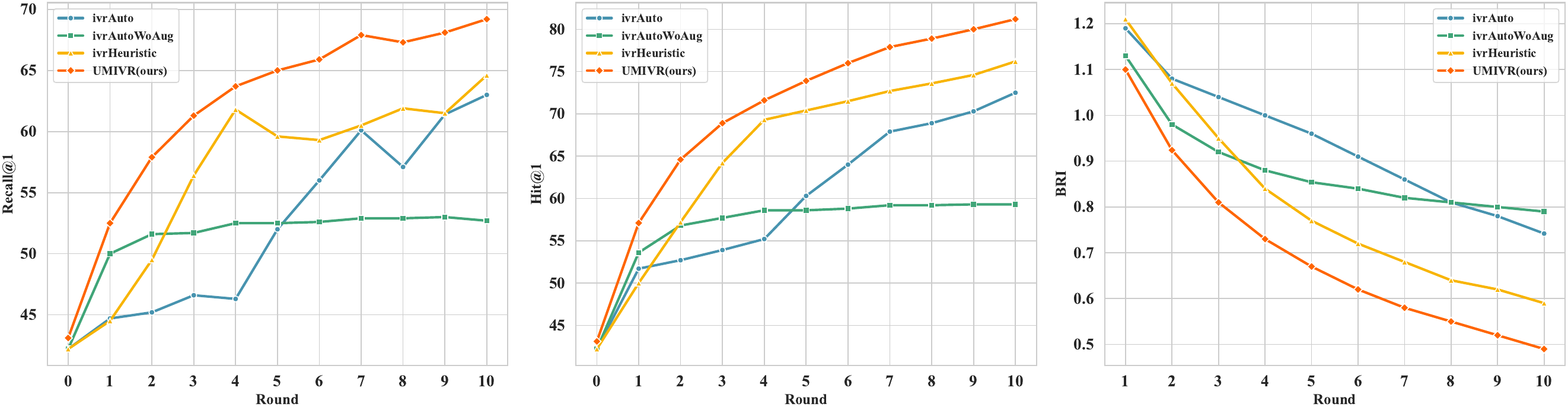}
    \caption{\textbf{Comparison of UMIVR with interactive baseline methods on MSR-VTT-1k across multiple interaction rounds.} From left to right, we illustrate Recall@1, Hit@1, and BRI scores, respectively. UMIVR consistently outperforms competing interactive baselines by achieving higher Recall@1 and Hit@1, as well as lower BRI values, highlighting its superior efficiency and effectiveness in iterative query refinement and retrieval accuracy improvement.}
    \label{fig:comparison}
\end{figure*}

\section{Experiments}

\subsection{Dataset and Evaluation}

\minisection{Datasets}
We perform experiments on four established TVR datasets: \textbf{MSR-VTT}~\cite{xu2016msr-vtt}, featuring 10,000 short videos each with 20 captions, using the standard 1K evaluation split (we also conduct our ablation studies on this dataset); \textbf{AVSD}~\cite{avsd}, providing dialogues grounded in videos, evaluated on a standard 1,000-sample test subset following prior works~\cite{ivr_base,d2v,vired}; \textbf{MSVD}~\cite{msvd}, consisting of around 2,000 videos annotated by multilingual captions, evaluated using the widely-adopted 670-video test subset; and \textbf{ActivityNet}~\cite{caba2015activitynet}, a large-scale dataset of untrimmed videos capturing 200 activity categories, evaluated on the commonly-used validation set (4,917 videos)~\cite{eercf, clipvip, Zhang2018CrossModalAH}.

\minisection{Evaluation Metrics}
We adopt three metrics for comprehensive evaluation: \textbf{Recall}, \textbf{Hit@\(k\)}, and the recent \textbf{Best log Rank Integral (BRI)}~\cite{plugir} metric. \textit{Recall} as a standard metric measures retrieval accuracy at fixed ranks. \textit{Hit} commonly used in interactive retrieval, reflects whether the target appears in the top-\(k\) candidates at any interaction step. 

BRI~\cite{plugir} is a new metric for interactive retrieval, integrating three essential aspects: (1) \emph{user satisfaction}, checking if the target video is eventually retrieved; (2) \emph{retrieval efficiency}, encouraging successful retrieval with fewer interactions; and (3) \emph{ranking improvement significance}, rewarding substantial improvements at higher ranks. By synthesizing these factors into one unified score, BRI aligns better with real user interaction scenarios, enabling nuanced performance comparisons among interactive retrieval methods.

\begin{table}[tbp]
\centering
\resizebox{0.5\textwidth}{!}
{
\begin{tabular}{lcccccc}
\toprule[1pt]
\diagbox{Methods}{Metrics} & R@1 $\uparrow$ & R@5 $\uparrow$ & R@10 $\uparrow$ & MnR $\downarrow$ & Hit@1 $\uparrow$ & Hit@10 $\uparrow$ \\ \hline
\multicolumn{7}{c}{\textit{Non-interactive TVR}} \\ \hline
CLIP4Clip \cite{clip4clip} & 44.5 & 71.4 & 81.6 & 15.3 & 44.5 & 81.6 \\
ProST \cite{prost} & 49.5 & 75.0 & 84.0 & 11.7 & 49.5 & 84.0 \\
UCOFIA \cite{ucofia} & 49.4 & 72.1 & - & - & 49.4 & - \\
EERCF \cite{eercf} & 54.1 & 78.8 & 86.9 & - & 54.1 & 86.9 \\
CLIP-ViP \cite{clipvip} & 57.7 & 80.5 & 88.2 & - & 57.7 & 88.2 \\
HunYuan(SOTA) \cite{hunyuan} & 62.9 & 84.5 & 90.8 & 9.3 & 62.9 & 90.8 \\ \hline
\multicolumn{7}{c}{\textit{UMIVR (Interactive TVR)}} \\ \hline
\textit{round 0} & 43.1 & 66.1 & 75.8 & 22.4 & 43.1 & 75.8 \\
\textit{round 2} & 57.9 & 81.2 & 86.6 & 10.4 & 57.2 & 89.9 \\
\textit{round 3} & 61.3 & 84.1 & 89.0 & {\ul 8.1} & {\ul 68.9} & {\ul 92.7} \\
\textit{round 6} & {\ul 65.9} & {\ul 87.7} & {\ul 91.8} & 5.9 & 76.0 & 95.3 \\
\textit{round 8} & \textbf{67.3} & \textbf{88.3} & \textbf{92.8} & \textbf{5.7} & \textbf{78.9} & \textbf{96.5} \\ \bottomrule[1pt]
\end{tabular}
}
\caption{\textbf{Comparison with non-interactive TVR methods on MSR-VTT-1k.} UMIVR significantly improves retrieval performance through multiple rounds of interaction. Notably, after at most three rounds of interaction, UMIVR surpasses most competing methods. Underlined values indicate the earliest interaction round where UMIVR exceeds existing approaches, while bold values denote the best overall performance across all methods.}
\label{tab:compare_nonivr}
\end{table}

\begin{table}[tbp]
\centering
\resizebox{0.5\textwidth}{!}
{
\begin{tabular}{lcccccc}
\toprule[1pt]
Methods & Round & R@1 $\uparrow$ & R@10 $\uparrow$ & Hit@1 $\uparrow$ & Hit@10 $\uparrow$ & BRI $\downarrow$ \\ \hline
\multirow{4}{*}{{D2V\cite{d2v}}} & 0 & 8.8 & 32.1 & 8.8 & 32.1 & - \\
 & 2 & 22.9 & 54.0 & - & - & - \\
 & 4 & 22.5 & 58.5 & - & - & - \\
 & 6 & 23.9 & 61.0 & - & - & - \\ \hline
VIRED \cite{vired} & 3 & 24.9 & 60.8 & - & - & - \\ \hline
\multirow{4}{*}{{ivrAuto}} & 0 & 29.6 & 61.3 & 29.6 & 61.3 & - \\
 & 2 & 30.7 & 62.2 & 35.9 & 66.9 & 1.74 \\
 & 4 & 32.6 & 66.7 & 39.4 & 71.6 & 1.67 \\
 & 6 & 37.3 & 74.3 & 45.1 & 78.9 & 1.57 \\ \hline
\multirow{4}{*}{{ivrAutoWoAug}} & 0 & 29.6 & 61.3 & 29.6 & 61.3 & - \\
 & 2 & 34.6 & 67.2 & 39.4 & 71.6 & 1.64 \\
 & 4 & 34.6 & 67.3 & 40.4 & 72.6 & 1.56 \\
 & 6 & 34.6 & 67.2 & 40.8 & 72.8 & 1.53 \\ \hline
\multirow{4}{*}{{ivrHeuristic}} & 0 & 29.6 & 61.3 & 29.6 & 61.3 & - \\
 & 2 & 35.2 & 69.2 & 41.6 & 73.5 & 1.61 \\
 & 4 & 43.2 & 80.3 & 51.1 & 82.7 & 1.38 \\
 & 6 & 43.4 & 80.8 & 53.5 & 86.2 & 1.23 \\ \hline
\multirow{4}{*}{{UMIVR(ours)}} & 0 & 30.6 & 61.6 & 30.6 & 61.6 & - \\
 & 2 & 44.8 & 78.9 & 50.8 & 81.5 & 1.40 \\
 & 4 & 47.9 & 81.4 & 58.8 & 86.7 & 1.16 \\
 & 6 & \textbf{49.9} & \textbf{82.2} & \textbf{63.3} & \textbf{88.2} & \textbf{1.02} \\ \bottomrule[1pt]
\end{tabular}
}
\caption{Comparison results on AVSD dataset.}
\label{tab:avsd}
\end{table}

\subsection{Implementation Details}
Our implementation is based on the open-source VideoLLaVA codebase\footnote{\href{https://github.com/PKU-YuanGroup/Video-LLaVA}{https://github.com/PKU-YuanGroup/Video-LLaVA}} with Python-3.10 and Pytorch-2.0.1. Specifically, we adopt VideoLLaVA-7B as our backbone model, leveraging 4-bit quantization to significantly reduce GPU memory consumption. To ensure consistency and reproducibility, we set the generation temperature to 0.1 for internal modules such as Captioner, Question Generation, and Query Refinement. Conversely, for the VideoQA module simulating diverse user responses, we set the generation temperature to 0.7. All other generation settings follow the default configurations provided by the codebase.

For visual encoding, we utilize the default LanguageBind encoder integrated within VideoLLaVA-7B, a post-pretrained CLIP ViT-L/14 model. This visual encoder processes RGB inputs with spatial dimensions of $224\times224$ and employs a patch size of 14 pixels, consisting of 24 transformer layers with temporal attention enabled. Each transformer layer features a hidden dimensionality of 1024 and 16 attention heads, providing high representational capability for both spatial and temporal information. The encoder supports video inputs of 8 frames and projects final vision embeddings to a dimensionality of 768.

\minisection{Baselines.} For baseline comparisons, we primarily compete with IVR~\cite{ivr_base}, a recent influential approach in interactive text-to-video retrieval that introduces an LLM-based training-free paradigm and demonstrates strong performance on MSR-VTT, MSVD, and AVSD datasets. IVR encompasses three variants: \textit{ivrHeuristic} (using manually-defined question templates), \textit{ivrAuto} (generating questions from top-k similar video captions assisted by heuristic-based augmentation), and \textit{ivrAutoWoAug} (identical to \textit{ivrAuto} but without heuristic augmentation). To ensure fair and rigorous comparisons, we faithfully reproduce these three IVR variants within our UMIVR framework, integrating the IVR codebase\footnote{\href{https://github.com/kevinliang888/IVR-QA-baselines}{https://github.com/kevinliang888/IVR-QA-baselines}} into the Video-LLaVA system. Additionally, since IVR's original implementation generates multiple questions concurrently without iterative interaction, we adapt it to strictly follow standard interactive retrieval conventions~\cite{plugir, merlin}, allowing only one question-answer exchange per interaction round. After this adaptation, our IVR baselines and UMIVR framework are fully aligned and directly comparable. Consistent with prior interactive retrieval studies and realistic application scenarios, we limit the maximum number of interaction rounds to 10, as exceeding this typically results in poor user experience.

\begin{table}[tbp]
\centering
\resizebox{0.5\textwidth}{!}
{
\begin{tabular}{ll|cccc|cccc}
\toprule[1pt]
\multirow{2}{*}{} & \multicolumn{1}{l|}{\multirow{2}{*}{\textbf{Methods}}} & \multicolumn{4}{c|}{\textbf{MSVD}} & \multicolumn{4}{c}{\textbf{ActivityNet}} \\
 & \multicolumn{1}{c|}{} & 0 & 1 & 3 & 5 & 0 & 1 & 3 & 5 \\ \hline
\multirow{4}{*}{\rotatebox{90}{\textbf{R@1}}} & ivrAuto & 50.3 & 52.8 & 54.4 & 61.7 & 32.9 & 33.8 & 33.9 & 35.1 \\
 & ivrAutowoAug & 50.3 & 55.5 & 58.9 & 59.2 & 32.9 & 36.3 & 36.4 & 36.5 \\
 & ivrHeuristic & 50.3 & 53.4 & 64.6 & 67.3 & 32.9 & 34.0 & 37.8 & 39.9 \\
 & UMIVR(ours) & \textbf{51.9} & \textbf{61.2} & \textbf{67.0} & \textbf{69.7} & \textbf{33.1} & \textbf{38.3} & \textbf{40.4} & \textbf{41.8} \\ \hline
\multirow{4}{*}{\rotatebox{90}{\textbf{R@10}}} & ivrAuto & 85.0 & 87.7 & 88.2 & 92.8 & 73.4 & 74.5 & 74.7 & 76.4 \\
 & ivrAutowoAug & 85.0 & 88.9 & 89.8 & 90.4 & 73.4 & 77.0 & 77.2 & 77.3 \\
 & ivrHeuristic & 85.0 & 87.1 & 90.5 & 91.3 & 73.4 & 76.1 & 79.2 & 80.5 \\
 & UMIVR(ours) & \textbf{86.4} & \textbf{90.9} & \textbf{93.9} & \textbf{94.8} & \textbf{73.7} & \textbf{78.2} & \textbf{80.0} & \textbf{80.7} \\ \hline
\multirow{4}{*}{\rotatebox{90}{\textbf{Hit@1}}} & ivrAuto & 50.3 & 57.7 & 59.8 & 68.5 & 32.9 & 36.9 & 37.3 & 39.7 \\
 & ivrAutowoAug & 50.3 & 61.0 & 65.6 & 66.1 & 32.9 & 39.2 & 40.5 & 40.6 \\
 & ivrHeuristic & 50.3 & 58.0 & 70.8 & 77.0 & 32.9 & 35.5 & 41.5 & 44.8 \\
 & UMIVR(ours) & \textbf{51.9} & \textbf{65.4} & \textbf{74.6} & \textbf{79.3} & \textbf{33.1} & \textbf{42.0} & \textbf{47.8} & \textbf{50.1} \\ \hline
\multirow{4}{*}{\rotatebox{90}{\textbf{Hit@10}}} & ivrAuto & 85.0 & 91.3 & 91.7 & 95.2 & 73.4 & 77.3 & 77.7 & 79.6 \\
 & ivrAutowoAug & 85.0 & 91.4 & 92.3 & 92.9 & 73.4 & 79.1 & 79.8 & 79.9 \\
 & ivrHeuristic & 85.0 & 89.5 & 93.1 & 94.5 & 73.4 & 77.2 & 81.3 & 83.0 \\
 & UMIVR(ours) & \textbf{86.4} & \textbf{92.5} & \textbf{95.5} & \textbf{96.7} & \textbf{73.7} & \textbf{79.9} & \textbf{83.0} & \textbf{84.1} \\ \hline
\multirow{4}{*}{\rotatebox{90}{\textbf{BRI}}} & ivrAuto & - & 0.85 & 0.75 & 0.68 & - & 1.40 & 1.35 & 1.32 \\
 & ivrAutowoAug & - & 0.82 & 0.68 & 0.64 & - & 1.37 & 1.28 & 1.26 \\
 & ivrHeuristic & - & 0.86 & 0.64 & 0.51 & - & 1.42 & 1.30 & 1.22 \\
 & UMIVR(ours) & - & \textbf{0.78} & \textbf{0.58} & \textbf{0.49} & - & \textbf{1.35} & \textbf{1.20} & \textbf{1.13} \\ \bottomrule[1pt]
\end{tabular}
}
\caption{Comparison results on MSVD and ActivityNet Dataset.}
\label{tab:msvd_acnet}
\end{table}

\subsection{Comparison Results}

Fig.~\ref{fig:comparison} and Tab.~\ref{tab:compare_nonivr} summarize retrieval performance on the MSR-VTT dataset. UMIVR consistently surpasses interactive baselines across interaction rounds, demonstrating clear advantages in retrieval accuracy and efficiency. Moreover, UMIVR quickly outperforms the leading non-interactive method (HunYuan\_tvr~\cite{hunyuan}). Specifically, after only three rounds, UMIVR already surpasses the HunYuan\_tvr in Hit@1 (68.9 vs. 62.9), highlighting its effectiveness in leveraging iterative interactions to explicitly address retrieval uncertainty.

Tab.~\ref{tab:avsd} and Tab.~\ref{tab:msvd_acnet} further confirm UMIVR's robust performance and generalization capabilities. Specifically, UMIVR exhibits substantial improvements over baselines on AVSD, MSVD, and ActivityNet datasets, maintaining consistent superiority across metrics and interaction rounds. These results collectively validate the effectiveness of explicitly quantifying and minimizing uncertainty in interactive text-to-video retrieval scenarios.

\begin{table}[tbp]
\centering
\resizebox{0.3\textwidth}{!}
{
\begin{tabular}{ll|ccc}
\toprule[1pt]
\multirow{3}{*}{\rotatebox{90}{Comp.}} & +\textit{TAS} & \ding{52} & \ding{52} & \ding{52} \\
 & +\textit{MUS} &  & \ding{52} & \ding{52} \\
 & +\textit{TQFS} &  &  & \ding{52} \\ \hline
\multirow{3}{*}{\rotatebox{90}{R@1}} & 1 & 51.6 & 52.2 & \textbf{52.5} \\
 & 3 & 61.0 & \textbf{62.1} & 61.3 \\
 & 5 & 63.4 & 64.2 & \textbf{65.0} \\ \hline
\multirow{3}{*}{\rotatebox{90}{Hit@1}} & 1 & 56.6 & \textbf{57.4} & 57.1 \\
 & 3 & 67.0 & 68.6 & \textbf{68.9} \\
 & 5 & 73.0 & 72.8 & \textbf{73.9} \\ \hline
\multirow{3}{*}{\rotatebox{90}{Hit@10}} & 1 & 86.1 & 86.4 & \textbf{86.7} \\
 & 3 & 91.1 & 92.5 & \textbf{92.7} \\
 & 5 & 93.7 & 94.4 & \textbf{94.8} \\ \hline
BRI & 5 & 0.69 & \textbf{0.67} & \textbf{0.67} \\ \bottomrule[1pt]
\end{tabular}
}
\caption{Ablation study on different components.}
\label{tab:comp}
\end{table}

\begin{table}[bp]
\centering
\resizebox{0.5\textwidth}{!}
{
\begin{tabular}{c|ccc|ccc|ccc|c}
\toprule[1.2pt]
\multirow{2}{*}{\textbf{($\alpha, \beta$)}} & \multicolumn{3}{c|}{R@1 $\uparrow$} & \multicolumn{3}{c|}{Hit@1 $\uparrow$} & \multicolumn{3}{c|}{Hit@10 $\uparrow$} & BRI $\downarrow$ \\
 & 1 & 3 & 5 & 1 & 3 & 5 & 1 & 3 & 5 & 5 \\ \hline
(0.4, 0.1) & 52.3 & 55.9 & 61.9 & 57.0 & 65.1 & 71.5 & 85.8 & 91.3 & 93.7 & 0.73 \\
(0.5, 0.1) & 51.9 & 57.9 & 63.1 & 56.4 & 66.1 & 72.4 & 86.2 & 92.6 & 94.6 & 0.70 \\
(0.6, 0.1) & 51.5 & 58.4 & 64.6 & 55.8 & 65.1 & 70.4 & 85.1 & 91.6 & \textbf{94.8} & 0.68 \\
(0.4, 0.2) & 52.5 & 60.4 & 63.9 & 57.0 & 67.6 & 73.4 & 86.2 & 91.3 & 93.9 & 0.69 \\
\textbf{(0.5, 0.2)} & 52.5 & \textbf{61.3} & \textbf{65.0} & \textbf{57.1} & 68.9 & \textbf{73.9} & \textbf{86.7} & 92.7 & \textbf{94.8} & \textbf{0.67} \\
(0.6, 0.2) & 51.9 & 61.1 & 64.7 & 56.6 & 68.5 & 73.1 & 85.9 & \textbf{92.9} & 94.5 & 0.68 \\
(0.4, 0.3) & 51.5 & 58.9 & 64.9 & 56.8 & 67.1 & 73.5 & 86.6 & 91.8 & 94.4 & 0.69 \\
(0.5, 0.3) & \textbf{52.6} & \textbf{61.3} & 64.6 & \textbf{57.1} & \textbf{70.0} & 73.6 & 86.0 & 92.8 & 94.4 & 0.68 \\
(0.6, 0.3) & 52.2 & 61.1 & 64.6 & 56.7 & 68.0 & 73.5 & 86.4 & 92.5 & 94.2 & 0.68 \\ \bottomrule[1.2pt]
\end{tabular}
}
\caption{Grid search for TAS threshold $\alpha$ and MUS threshold $\beta$.}
\label{tab:grid_search}
\end{table}

\subsection{Ablation Study}

We perform comprehensive ablation analyses to investigate the contribution of each uncertainty-related component in UMIVR and examine the sensitivity of threshold parameters.
Tab.~\ref{tab:comp} explores the individual impact of Text Ambiguity Score (TAS), Mapping Uncertainty Score (MUS), and Temporal Quality-based Frame Sampler (TQFS). Clearly, integrating each component incrementally improves retrieval performance, and their joint usage delivers the best results across metrics. Particularly, MUS and TAS demonstrate complementary roles, jointly addressing uncertainties in textual ambiguity and text-video mapping, while TQFS further boosts performance by reducing visual noise.

To understand parameter sensitivity, Tab.~\ref{tab:grid_search} presents grid search results for threshold parameters $\alpha$ (TAS) and $\beta$ (MUS). Optimal retrieval performance emerges at $(\alpha,\beta) = (0.5, 0.2)$, consistently achieving highest Recall@1, Hit@1, Hit@10, and lowest BRI. Deviating from these optimal values results in noticeable performance drops, validating the importance of jointly tuning these thresholds. These analyses confirm that UMIVR effectively leverages explicit uncertainty modeling, and carefully calibrated thresholds significantly enhance retrieval accuracy.

\begin{table}[tbp]
\centering
\resizebox{0.5\textwidth}{!}
{
\begin{tabular}{lccc|ccc}
\toprule[1pt]
 & \multicolumn{3}{c|}{\textit{Text-to-Video Retrieval}} & \multicolumn{3}{c}{\textit{Video-to-Text Retrieval}} \\
\textbf{Methods} & \textbf{R1} $\uparrow$ & \textbf{R10} $\uparrow$ & \textbf{MnR} $\downarrow$ & \textbf{R1} $\uparrow$ &  \textbf{R10} $\uparrow$ & \textbf{MnR} $\downarrow$ \\  \hline
VideoCLIP \cite{video-clip} & 30.7 & 71.0 & 18.1 & 30.2 & 70.4 & 23.0 \\
\cite{video-clip} + \textbf{TQFS} & { \textbf{31.1({\color{ForestGreen}+0.4})}} & \textbf{72.4({\color{ForestGreen}+1.4})} & \textbf{16.1({\color{ForestGreen}-2.0})} & \textbf{31.1({\color{ForestGreen}+0.9})} & \textbf{72.4({\color{ForestGreen}+2.0)}} & \textbf{21.0({\color{ForestGreen}-2.0})} \\
Xpool \cite{xpool} & 45.3 & 80.2 & 15.9 & 43.0 & \textbf{83.2} & 10.5 \\
\cite{xpool} + \textbf{TQFS} & \textbf{45.7({\color{ForestGreen}+0.4})} & \textbf{80.9({\color{ForestGreen}+0.7})} & \textbf{14.3({\color{ForestGreen}-1.6})} & \textbf{43.7(\color{ForestGreen}+0.7)} & 82.7(\color{Crimson}-0.5) & \textbf{9.2(\color{ForestGreen}-1.3)} \\
\bottomrule[1pt]
\end{tabular}
}
\caption{TQFS enhances performance as a plug-in module.}
\label{tab:gener_tqfs}
\end{table}

\begin{table}[bp]
\centering
\resizebox{0.5\textwidth}{!}
{
\begin{tabular}{llcccccc}
\toprule[1pt]
 & \textit{Round} & 0 & 2 & 4 & 6 & 8 & 10 \\ \hline
\multirow{4}{*}{\rotatebox{90}{\textbf{R@10}}} & \multicolumn{1}{l|}{ChatIR\cite{chatir}} & 71.5 & 74.9 & 76 & 78.2 & 78.8 & 79.5 \\
 & \multicolumn{1}{l|}{PlugIR\cite{plugir}} & 71.1 & 75.5 & 76.1 & 76.1 & 75.1 & 74.3 \\
 & \multicolumn{1}{l|}{ivrHeuristic} & \textbf{73.2} & 78.7 & 76.7 & 78.7 & 77.8 & 80.3 \\
 & \multicolumn{1}{l|}{UMIVR(ours)} & \textbf{73.2} & \textbf{83.0} & \textbf{83.6} & \textbf{84.5} & \textbf{85.1} & \textbf{85.0} \\ \hline
\multirow{4}{*}{\rotatebox{90}{\textbf{Hit@10}}} & \multicolumn{1}{l|}{ChatIR\cite{chatir}} & 71.5 & 78.9 & 82.0 & 84.4 & 85.6 & 86.4 \\
 & \multicolumn{1}{l|}{PlugIR\cite{plugir}} & 71.1 & 83.1 & 87.6 & 89.4 & 90.7 & 91.5 \\
 & \multicolumn{1}{l|}{ivrHeuristic} & \textbf{73.2} & 79.5 & 84.1 & 85.7 & 86.2 & 86.8 \\
 & \multicolumn{1}{l|}{UMIVR(ours)} & \textbf{73.2} & \textbf{87.0} & \textbf{89.1} & \textbf{90.4} & \textbf{91.3} & \textbf{91.8} \\ \bottomrule[1pt]
\end{tabular}
}
\caption{UMIVR achieves competitive improvements in interactive text-to-image retrieval on VisDial dataset.}
\label{tab:gener_visdial}
\end{table}

\subsection{Generalization}

We further investigate the generalization capability of UMIVR and its components across broader scenarios and modalities.
Tab.~\ref{tab:gener_tqfs} demonstrates that our Temporal Quality-based Frame Sampler (TQFS) effectively serves as a plug-and-play module. When directly integrated into trained single-round TVR methods (VideoCLIP~\cite{video-clip} and Xpool~\cite{xpool}), TQFS achieves consistent and meaningful improvements on MSR-VTT-1ka (e.g., gains of +0.4\% to +1.4\% in Recall@10 and reductions in MnR), confirming its versatility and effectiveness in enhancing existing retrieval methods without additional fine-tuning.

Furthermore, leveraging the multimodal capabilities of VideoLLaVA and LanguageBind, we extend UMIVR to interactive text-to-image retrieval. We evaluate UMIVR on the Visual Dialog (VisDial) \cite{visdial} dataset, which includes 8,000 test images and corresponding captions from human-annotated dialogues based on MS-COCO~\cite{mscoco}. As shown in Tab.~\ref{tab:gener_visdial}, UMIVR consistently outperforms competitive interactive retrieval methods (ChatIR~\cite{chatir}, PlugIR~\cite{plugir}, and ivrHeuristic) across interaction rounds. Specifically, UMIVR achieves noticeable improvements in Recall@10 and Hit@10, highlighting its robustness and generalizability beyond video retrieval tasks.

\begin{figure}[tbp]
    \centering
    \includegraphics[width=0.48\textwidth]{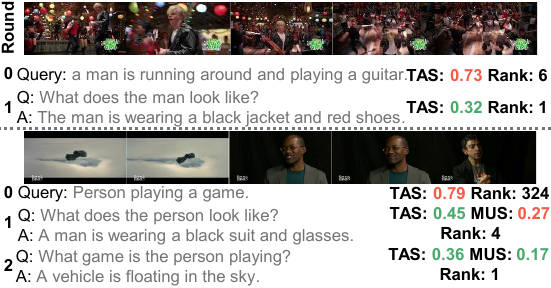}
    \caption{\textbf{Case study of UMIVR’s interactive retrieval process.} The examples illustrate how uncertainty-aware question generation progressively refines ambiguous queries, reducing the Text Ambiguity Score (TAS) and Mapping Uncertainty Score (MUS) while improving retrieval rank.}
    \label{fig:case_study}
\end{figure} 

\subsection{Case Study}

Fig.~\ref{fig:case_study} illustrates two qualitative examples demonstrating UMIVR's interactive retrieval process. Initially, queries contain significant ambiguity, resulting in high TAS and low rankings. Through uncertainty-minimizing clarifying questions, UMIVR progressively refines the queries by explicitly addressing textual and mapping uncertainties. This refinement effectively reduces TAS and MUS, substantially improving retrieval ranks after each interaction round. These examples highlight UMIVR's effectiveness in adaptively resolving uncertainty to achieve precise retrieval.

\section{Conclusion}

In this paper, we introduced UMIVR, an uncertainty-aware interactive framework for text-to-video retrieval that systematically quantifies and minimizes three fundamental uncertainties—text ambiguity, mapping uncertainty, and frame uncertainty. By proposing principled, training-free uncertainty metrics, UMIVR adaptively generates clarifying questions, iteratively refining user intent and significantly enhancing retrieval accuracy. Extensive experiments on benchmarks including MSR-VTT-1k demonstrated that UMIVR surpasses prior interactive and non-interactive methods, highlighting its effectiveness and generalizability. Our work thus establishes a robust uncertainty-minimizing foundation for interactive multimodal retrieval, opening promising directions for future research in uncertainty-aware interactive learning across vision-language tasks.

\clearpage

\section*{Acknowledgements}
 This work is supported by Australian Research Council (ARC) Discovery Project DP230101753, and CSIRO’s Science Leader Project R-91559. 


{
    \small
    \bibliographystyle{ieeenat_fullname}
    \bibliography{main}
}

\clearpage
\onecolumn
\appendix 

\appendixpage
\addappheadtotoc

\noindent\textbf{Appendix Contents}\par
\begin{enumerate}
  \item Explanation of UMIVR \dotfill \pageref{sec:explanation}
  \item Reproduction of IVR Baselines \dotfill \pageref{sec:repro-ivr}
  \item Analysis of Uncertainty Score Distributions \dotfill \pageref{sec:uncertainty-dist}
  \item Discussion on Quantization Strategies \dotfill \pageref{sec:quant}
  \item Early Stopping Strategy for Interactive Retrieval \dotfill \pageref{sec:early-stop}
  \item Impact of NR-IQA Methods in TQFS \dotfill \pageref{sec:impact-nriqa}
  \item Prompt Design for UMIVR \dotfill \pageref{sec:prompt-design}
  \item Limitations \dotfill \pageref{sec:limitations}
\end{enumerate}

\section{Explanation of UMIVR}
\label{sec:explanation}

Uncertainty Reduction Theory (URT)~\cite{berger1974some}, originates from interpersonal communication studies and addresses how individuals seek information during interactions to alleviate uncertainty and enhance predictability. According to URT, uncertainty arises when communicators cannot accurately predict outcomes or interpret messages due to incomplete or ambiguous information. URT emphasizes that this uncertainty negatively impacts interactions, motivating communicators to actively gather additional information through questions, clarifications, or direct observations. Mathematically, the concept of uncertainty in URT aligns closely with Shannon's entropy in information theory, where greater entropy indicates higher unpredictability.

In interactive TVR, similar uncertainty challenges exist due to ambiguity in user queries and variability in the visual content of videos. 
Our UMIVR framework aligns naturally with URT principles by explicitly quantifying uncertainty through well-defined measures—semantic entropy~\cite{seman_entropy_iclr,seman_entropy_nat} for text ambiguity, Jensen–Shannon divergence for mapping ambiguity, and quality-based sampling for frame-level ambiguity—and systematically reducing these uncertainties via iterative interactions. Specifically, by leveraging these principled information-theoretic measures, UMIVR embodies URT’s active information-seeking approach, enabling progressively clearer and more accurate text-to-video retrieval outcomes.

\section{Reproduction of IVR Baselines}
\label{sec:repro-ivr}

To ensure a rigorous and fair comparison with the proposed UMIVR approach, we faithfully reproduced the IVR baseline methods \cite{ivr_base} within our experimental setting. This section elaborates on the implementation specifics, evaluation outcomes, and computational considerations of baselines.

\minisection{Baseline Implementation}
We re-implement two variants of IVR: \textit{ivrAuto}, which employs large language models for automatic clarifying question generation, and \textit{ivrHeuristic}, which utilizes heuristic-based strategies for question generation. Besides, \textit{ivrHeuristicWoAug} can be simply implemented by changing the config files. To align IVR variants closely with our proposed UMIVR framework, we re-implemented both variants using the unified multimodal model, VideoLLaVA. Specifically, the original ensemble-based components were replaced with the 4-bit quantized VideoLLaVA-7B model. Hyperparameters and experimental conditions were matched closely with the original setups to maintain fairness in comparisons.

\begin{figure*}[tbp]
    \centering
    \includegraphics[width=0.9\textwidth]{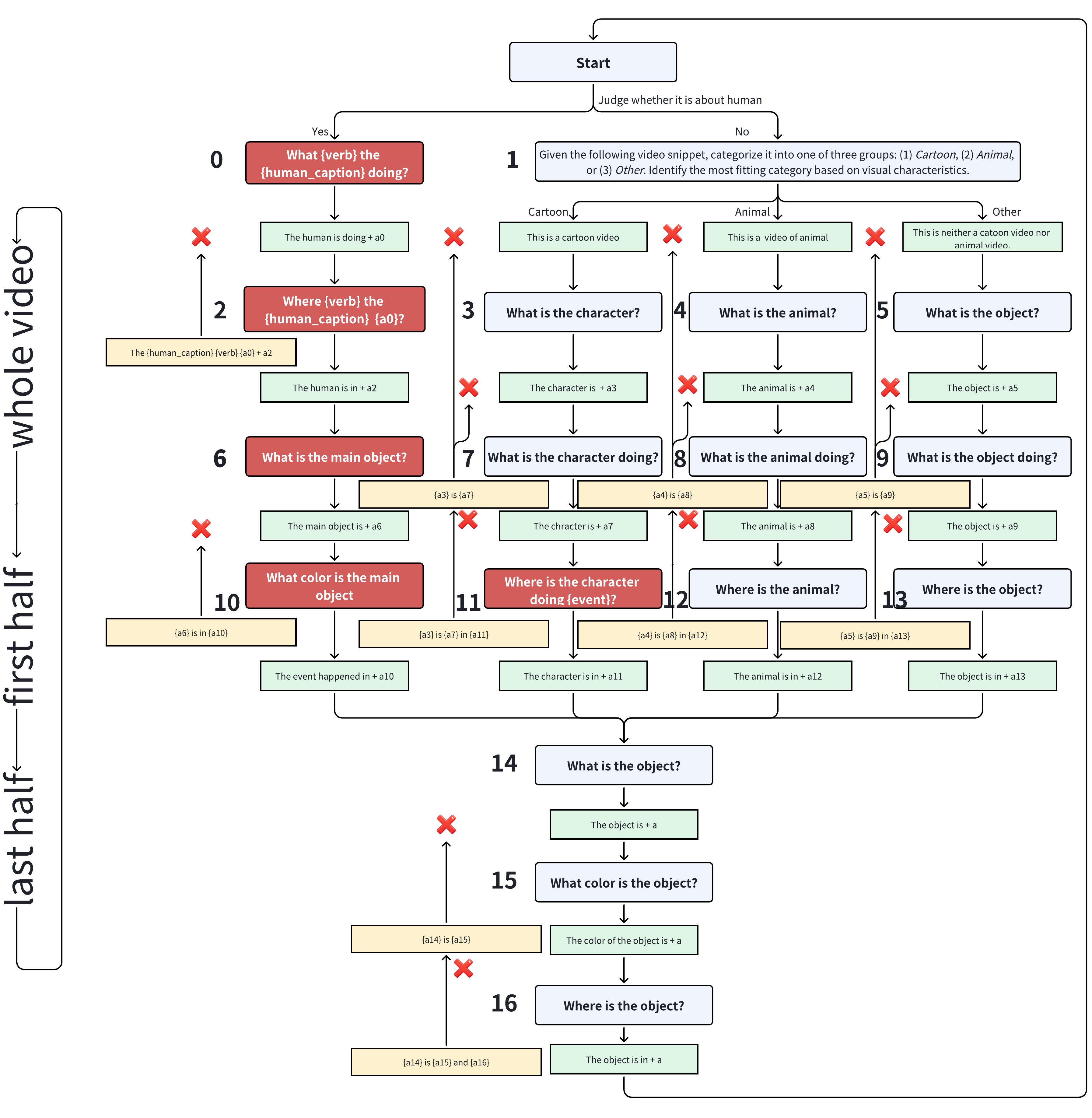}
    \caption{\textbf{Overview of the Reproduced IVR-Heuristic Pipeline Structure.} The pipeline follows a structured, heuristic-driven interactive retrieval approach for text-to-video retrieval. It first categorizes videos into four types: human, cartoon, animal, and other, tailoring its questioning strategy accordingly to extract key video features. To improve retrieval effectiveness, the pipeline segments video interactions into three stages: whole video, first half, and last half, enabling more granular query refinement. Additionally, to prevent missing critical details, it incorporates general questions at the end of each round, covering main objects, colors, and locations. The process iterates through multiple rounds until reaching a predefined maximum, progressively refining the query and enhancing retrieval accuracy.}
    \label{fig:ivr_heuristic_pipeline}
\end{figure*}

Figure~\ref{fig:ivr_heuristic_pipeline} provides a detailed illustration of the reproduced IVR heuristic pipeline. The pipeline systematically organizes interactions into iterative stages, initially focusing on the entire video, followed by granular interactions targeting the first and second halves separately. Additionally, general questions covering key objects, colors, and locations are integrated at the end of each interaction round, ensuring comprehensive query refinement. This structured reproduction closely adheres to the original IVR approach, effectively utilizing the capabilities of the VideoLLaVA.

\minisection{Results Analysis}
Table~\ref{tab:reproduce_res} presents comprehensive performance comparisons between our VideoLLaVA-based IVR reproductions and the original IVR implementations across multiple evaluation metrics on the MSR-VTT dataset. Our reproduced models demonstrate clear improvements, especially in later interaction rounds. Notably, the \textit{ivrHeuristic (VideoLLaVA)} model achieves significant gains across Hit@1, Hit@5, and Hit@10 metrics, confirming that leveraging a unified multimodal model like VideoLLaVA enhances query refinement and overall retrieval accuracy.

We also examine GPU memory consumption for various IVR implementations, as illustrated in Table~\ref{tab:gpu_mem}. Original IVRAuto implementations incur substantial memory usage, primarily due to their reliance on multiple distinct models (e.g., T0++). By contrast, integrating the unified VideoLLaVA architecture with 4-bit quantization significantly reduces GPU memory usage by nearly sixfold compared to original implementations (e.g., 9,196 MB vs. 59,451 MB for \textit{ivrAuto}), thereby greatly enhancing computational efficiency and practical deployment viability.

Overall, these reproduction efforts validate the robustness and effectiveness of employing a unified multimodal architecture, offering precise benchmarks for rigorous evaluation of the UMIVR framework.

\begin{table*}[htbp]
\centering
\begin{tabular}{lc|ccccccccccc}
\toprule[1pt]
\textbf{} & Rounds & 0 & 1 & 2 & 3 & 4 & 5 & 6 & 7 & 8 & 9 & 10 \\ \hline
\multirow{4}{*}{\textbf{Hit@1}} & \begin{tabular}[c]{@{}c@{}}ivrAuto\\ (original)\end{tabular} & 42.5 & 50.1 & 54.2 & 56.4 & 58.4 & 61.2 & 62.4 & 64.0 & 64.8 & 66.4 & 67.5 \\
 & \begin{tabular}[c]{@{}c@{}}ivrAuto\\ (VideoLLaVA)\end{tabular} & 42.2 & 51.7 & 52.7 & 53.9 & 55.2 & 60.3 & 64.0 & 67.9 & 68.9 & 70.3 & 72.5 \\
 & \begin{tabular}[c]{@{}c@{}}ivrHeuristic\\ (original)\end{tabular} & 42.5 & 49.0 & 56.0 & 63.0 & 67.2 & 68.9 & 71.5 & 73.0 & 74.0 & 74.9 & 75.5 \\
 & \begin{tabular}[c]{@{}c@{}}ivrHeuristic\\ (VideoLLaVA)\end{tabular} & 42.2 & 50.0 & 57.2 & 64.2 & 69.3 & 70.4 & 71.5 & 72.7 & 73.6 & 74.6 & 76.2 \\ \hline
\multirow{4}{*}{\textbf{Hit@5}} & \begin{tabular}[c]{@{}c@{}}ivrAuto\\ (original)\end{tabular} & 65.3 & 73.9 & 81.1 & 83.4 & 84.2 & 84.9 & 85.5 & 85.9 & 86.0 & 86.1 & 86.3 \\
 & \begin{tabular}[c]{@{}c@{}}ivrAuto\\ (VideoLLaVA)\end{tabular} & 65.8 & 75.6 & 76.6 & 76.9 & 79.2 & 83.2 & 87.2 & 89.0 & 89.3 & 90.2 & 91.2 \\
 & \begin{tabular}[c]{@{}c@{}}ivrHeuristic\\ (original)\end{tabular} & 65.3 & 72.0 & 78.9 & 84.3 & 86.3 & 88.3 & 89.3 & 90.7 & 91.4 & 91.7 & 92.2 \\
 & \begin{tabular}[c]{@{}c@{}}ivrHeuristic\\ (VideoLLaVA)\end{tabular} & 65.8 & 73.2 & 79.9 & 86.9 & 89.2 & 89.8 & 90.2 & 90.9 & 91.0 & 91.7 & 92.4 \\ \hline
\multirow{4}{*}{\textbf{Hit@10}} & \begin{tabular}[c]{@{}c@{}}ivrAuto\\ (original)\end{tabular} & 74 & 81.7 & 83.1 & 84.1 & 86.0 & 88.6 & 90.1 & 90.8 & 91.1 & 91.5 & 91.9 \\
 & \begin{tabular}[c]{@{}c@{}}ivrAuto\\ (VideoLLaVA)\end{tabular} & 75.3 & 83.9 & 84.4 & 84.8 & 87.5 & 90.3 & 92.2 & 93.3 & 93.3 & 93.7 & 94.3 \\
 & \begin{tabular}[c]{@{}c@{}}ivrHeuristic\\ (original)\end{tabular} & 74 & 81.1 & 86.8 & 90.5 & 92.0 & 93.2 & 93.8 & 94.7 & 95.1 & 95.1 & 95.8 \\
 & \begin{tabular}[c]{@{}c@{}}ivrHeuristic\\ (VideoLLaVA)\end{tabular} & 75.3 & 82.3 & 84.9 & 87.6 & 90.1 & 90.6 & 91.8 & 93.0 & 94.1 & 96.2 & 96.5 \\ \bottomrule[1pt]
\end{tabular}
\caption{\textbf{Reproduction results of IVR~\cite{ivr_base} with VideoLLaVA.} This table compares the original IVR method with its VideoLLaVA-based implementation across multiple interaction rounds. We evaluate two IVR variants—\textit{ivrAuto} and \textit{ivrHeuristic}—on standard retrieval metrics (Hit@1, Hit@5, and Hit@10). The results demonstrate that integrating VideoLLaVA into IVR leads to improved retrieval performance, particularly in later interaction rounds. }
\label{tab:reproduce_res}
\end{table*}

\begin{table*}[htbp]
\resizebox{1.0\textwidth}{!}
{
\begin{tabular}{l|ccccccc}
\toprule[1pt]
Model & \begin{tabular}[c]{@{}c@{}}ivrAuto\\ (original)\end{tabular} & \begin{tabular}[c]{@{}c@{}}ivrAuto(4bit)\\ (VideoLLaVA)\end{tabular} & \begin{tabular}[c]{@{}c@{}}ivrHeuristic\\ (original)\end{tabular} & \begin{tabular}[c]{@{}c@{}}ivrHeuristic(4bit)\\ (VideoLLaVA)\end{tabular} & \begin{tabular}[c]{@{}c@{}}UMIVR\\ (4bit)\end{tabular} & \begin{tabular}[c]{@{}c@{}}UMIVR\\ (8bit)\end{tabular} & \begin{tabular}[c]{@{}c@{}}UMIVT\\ (woquant)\end{tabular} \\ \hline
GPU-Memory-Usage(MB) & 59451 & 9196 & 14934 & 10010 & 10210 & 15605 & 19297 \\ \bottomrule[1pt]
\end{tabular}
}
\caption{GPU memory usage comparison of different models. We evaluate the GPU memory consumption of various interactive retrieval models on a subset of MSR-VTT-1kA (first 10 samples) during inference. The results show that \textbf{ivrAuto (original)} exhibits the highest memory usage due to its reliance on T0++\cite{t0} for captioning, significantly increasing computational overhead. In contrast, the \textbf{VideoLLaVA-based 4-bit quantized models} (ivrAuto (4bit), ivrHeuristic (4bit), and UMIVR (4bit)) achieve substantial memory savings while maintaining competitive performance, making them efficient alternatives. The proposed \textbf{UMIVR framework} effectively balances performance and memory efficiency, with its 4-bit quantized version consuming only a fraction of the GPU memory required by non-quantized baselines.}
\label{tab:gpu_mem}
\end{table*}

\section{Analysis of Uncertainty Score Distributions}
\label{sec:uncertainty-dist}

In this section, we conduct a detailed analysis of the distributions of the two uncertainty metrics introduced in our proposed UMIVR framework: the Text Ambiguity Score (TAS) and the Mapping Uncertainty Score (MUS). These analyses highlight the effectiveness and interpretability of our uncertainty metrics across iterative  rounds.

\begin{figure*}[tbp]
    \centering
    \includegraphics[width=1.0\textwidth]{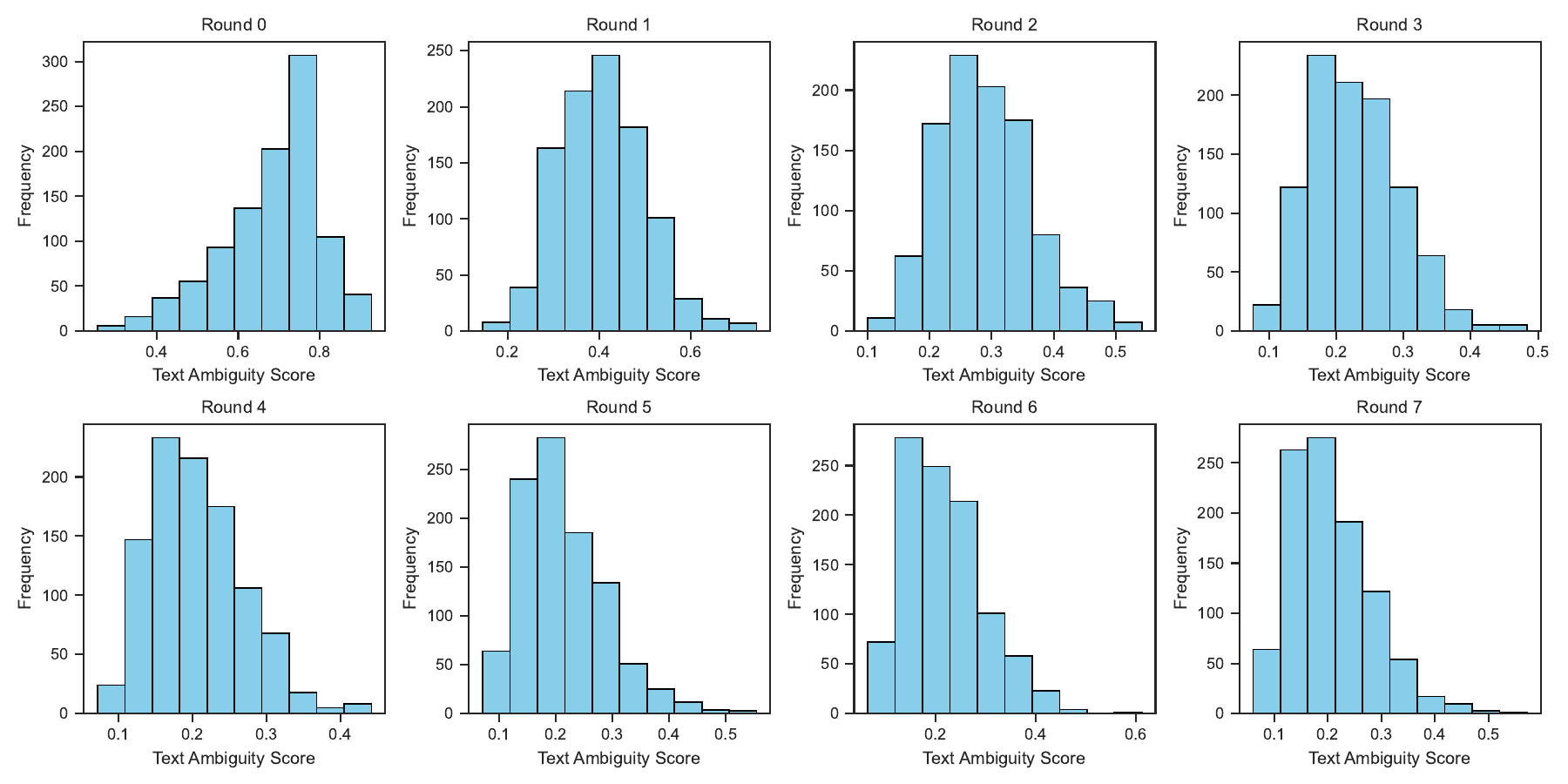}
    \caption{ \textbf{Distribution of Text Ambiguity Score (TAS) across interaction rounds.} The histograms illustrate the progressive reduction in TAS as the interactive retrieval process advances from Round 0 to Round 7. Initially, the majority of queries exhibit high ambiguity, with a strong concentration above 0.6 in Round 0. As clarifying questions iteratively refine the queries, the TAS distribution shifts leftward, indicating reduced textual ambiguity. By Round 3–5, most queries fall below the TAS threshold of 0.5 (marked as our uncertainty resolution threshold), and by Round 7, ambiguity is significantly minimized, demonstrating the effectiveness of UMIVR’s adaptive clarification strategy in refining textual queries.}
    \label{fig:tas_dis}
\end{figure*}

Figure~\ref{fig:tas_dis} shows the distributions of TAS values across multiple interaction rounds during the interactive retrieval process. At the initial stage (Round 0), queries exhibit notably high textual ambiguity, predominantly concentrated above a TAS value of 0.6. As the interactive retrieval proceeds and clarifying questions iteratively refine user queries, the TAS distributions consistently shift leftward, signaling a clear reduction in ambiguity. Specifically, from Rounds 3 to 5, the majority of queries already fall below our defined TAS uncertainty threshold of 0.5, indicating substantial resolution of textual ambiguity. By Round 7, the distribution further tightens around lower TAS values, underscoring the robustness of UMIVR's adaptive clarification strategy in progressively refining textual queries and reducing ambiguity.

\section{Discussion on Quantization Strategies}
\label{sec:quant}

We evaluate the effectiveness and computational trade-offs of different quantization strategies in UMIVR, comparing three configurations: UMIVR with 4-bit quantization (UMIVR-4bit), 8-bit quantization (UMIVR-8bit), and without quantization (UMIVR-woquant).

Table~\ref{tab:quant} reports retrieval performance across multiple interaction rounds. UMIVR-woquant consistently achieves the best results, particularly in later rounds, highlighting the advantages of full-precision models. However, the improvements over quantized versions remain modest. UMIVR-8bit slightly outperforms UMIVR-4bit, though the latter remains highly competitive with minimal degradation.

Table~\ref{tab:gpu_mem} shows GPU memory usage on the MSR-VTT dataset. Traditional IVR models incur high memory overhead due to ensemble-based architectures, whereas VideoLLaVA-based UMIVR significantly reduces memory consumption. Notably, UMIVR-4bit requires only 10,210 MB, less than half of UMIVR-8bit and one-third of UMIVR-woquant, demonstrating substantial efficiency.

Given the trade-off between accuracy and computational cost, we adopt UMIVR-4bit as the final configuration, offering near-optimal performance while drastically reducing GPU memory consumption, making it well-suited for real-world deployment.

\begin{table*}[htbp]
\centering
\begin{tabular}{llcccccc}
\toprule[1pt]
 & Rounds & 0 & 1 & 2 & 3 & 4 & 5 \\ \hline
\multirow{3}{*}{R@1} & \multicolumn{1}{l|}{UMIVR-4bit} & 43.1 & 52.5 & 57.9 & 61.3 & 63.7 & 65.0 \\
 & \multicolumn{1}{l|}{UMIVR-8bit} & 43.1 & 53.7 & 58.8 & 61.6 & 64.0 & 65.1 \\
 & \multicolumn{1}{l|}{UMIVR-woquant} & \textbf{43.8} & \textbf{54.2} & \textbf{58.9} & \textbf{63.3} & \textbf{65.4} & \textbf{67.1} \\ \hline
\multirow{3}{*}{Hit@10} & \multicolumn{1}{l|}{UMIVR-4bit} & 75.8 & 86.7 & 89.9 & 92.7 & 93.8 & 94.8 \\
 & \multicolumn{1}{l|}{UMIVR-8bit} & 75.8 & \textbf{87.6} & 91.6 & \textbf{94.3} & 95.2 & \textbf{95.9} \\
 & \multicolumn{1}{l|}{UMIVR-woquant} & \multicolumn{1}{r}{\textbf{76.3}} & 87.4 & \textbf{92.1} & 93.8 & \textbf{95.6} & 95.8 \\ \hline
\multirow{3}{*}{BRI} & \multicolumn{1}{l|}{UMIVR-4bit} & - & 1.10 & 0.92 & 0.81 & 0.73 & 0.67 \\
 & \multicolumn{1}{l|}{UMIVR-8bit} & - & 1.10 & \textbf{0.89} & \textbf{0.77} & \textbf{0.69} & 0.63 \\
 & \multicolumn{1}{l|}{UMIVR-woquant} & - & \textbf{1.09} & 0.90 & 0.78 & \textbf{0.69} & \textbf{0.62} \\ \bottomrule[1pt]
\end{tabular}
\caption{\textbf{Performance comparison of UMIVR with different quantization strategies.} We evaluate UMIVR under three settings: 4-bit quantization (UMIVR-4bit), 8-bit quantization (UMIVR-8bit), and non-quantized (UMIVR-woquant) across multiple interaction rounds. The results indicate that while UMIVR-woquant achieves the best overall performance, especially in later interaction rounds. However, the non-quantized version significantly increases GPU memory consumption, making it less practical. Meanwhile, UMIVR-8bit offers slightly better performance than UMIVR-4bit but at the cost of higher GPU usage. Given the trade-off between computational efficiency and retrieval performance, we adopt 4-bit quantization as the final configuration, as it achieves near-optimal results while maintaining significantly lower memory requirements.}
\label{tab:quant}
\end{table*}

\begin{table*}[htbp]
\centering
\begin{tabular}{l|ccccccccc}
\toprule[1pt]
 & Hit@1 & Hit@5 & Hit@10 & Rounds Min & Max & Mean & Median & 25\% & 75\% \\ \hline
UMIVR-EarlyStop & 65.8 & 84.7 & 90.7 & 1 & 10 & 3.05 & 3.0 & 2.0 & 3.0 \\ \bottomrule[1pt]
\end{tabular}
\caption{\textbf{Evaluation of UMIVR with an automatic early stopping mechanism.} Traditional interactive TVR methods typically run for a fixed number of interaction rounds (e.g., \( n = 10 \)), but real-world user interactions require a more dynamic approach to optimize user experience. To achieve this, we set a Text Ambiguity Score (TAS) threshold \( \alpha = 0.4 \) and a Mapping Uncertainty Score (MUS) threshold \( \beta = 0.2 \), allowing the system to automatically terminate interactions when the query is sufficiently refined. The results show that UMIVR-EarlyStop maintains strong retrieval performance while significantly reducing the average number of interaction rounds (\textbf{Mean = 3.05, Median = 3.0}). This validates the effectiveness of TAS and MUS as uncertainty indicators and demonstrates the practical benefits of adaptive interaction termination in improving efficiency and user experience.}
\label{tab:earlystop}
\end{table*}

\begin{table*}[b]
\centering
\begin{tabular}{l|cccccc}
\toprule[1pt]
NR-IQA Methods & R@1 & R@5 & R@10 & MdR & MnR & Runtime (s/video) \\ \hline
BRISQUE~\cite{brisque} & 43.3 & 66.4 & 75.9 & 2 & 22.3 & 4.09 \\
Laplacian Variance~\cite{petrou2010image}& 43.1 & 66.1 & 75.8 & 2 & 22.8 & 0.29 \\ \bottomrule[1pt]
\end{tabular}
\caption{\textbf{Comparison of TQFS with different No-Reference Image Quality Assessment (NR-IQA) methods.} We evaluate the impact of different NR-IQA methods on retrieval performance and runtime efficiency. While BRISQUE achieves slightly better results, it comes at a significant computational cost, requiring 4.09 seconds per video compared to 0.29 seconds for Laplacian Variance. Given the minimal performance difference but substantial speed advantage, we adopt Laplacian Variance in TQFS to ensure computational efficiency without sacrificing retrieval effectiveness.}
\label{tab:nriqa}
\end{table*}

\section{Early Stopping Strategy for Interactive Retrieval}
\label{sec:early-stop}

In practical interactive retrieval scenarios, optimizing the number of user interactions is crucial for enhancing user experience and efficiency. Traditional interactive text-to-video retrieval (TVR) methods often rely on a fixed number of interaction rounds, potentially leading to unnecessary or redundant interactions. To address this, we introduce an automatic early stopping strategy into our UMIVR framework, utilizing uncertainty metrics—specifically, the Text Ambiguity Score (TAS) and the Mapping Uncertainty Score (MUS)—as termination criteria.

As shown in Table~\ref{tab:earlystop}, by setting thresholds of \(\alpha = 0.4\) for TAS and \(\beta = 0.2\) for MUS, UMIVR effectively determines when further interactions become unnecessary. This adaptive early stopping mechanism achieves competitive retrieval performance (Hit\@1 = 65.8\%, Hit\@5 = 84.7\%, Hit\@10 = 90.7\%) while significantly reducing the average interaction rounds to approximately 3.05 (median of 3 rounds). The distribution of interaction rounds, ranging from a minimum of 1 to a maximum of 10, underscores the system's flexibility in dynamically adapting to varying query difficulties.

These results validate the practical effectiveness of using TAS and MUS as uncertainty-driven early stopping indicators, highlighting substantial improvements in retrieval efficiency and user interaction quality.

\section{Impact of NR-IQA Methods in TQFS}
\label{sec:impact-nriqa}

The Temporal Quality-based Frame Sampler (TQFS) introduced in our UMIVR framework relies on No-Reference Image Quality Assessment (NR-IQA) methods to select high-quality video frames effectively. Here, we briefly examine the impact of two NR-IQA methods—BRISQUE~\cite{brisque} and Laplacian Variance~\cite{petrou2010image}—on retrieval performance and runtime efficiency.

Table~\ref{tab:nriqa} compares the retrieval performance and computational costs associated with these NR-IQA methods. Although BRISQUE achieves marginally superior performance, it incurs significantly higher computational overhead (4.09 s/video) compared to the simpler Laplacian Variance method (0.29 s/video). Given this minimal difference in retrieval accuracy and substantial computational advantage, we select Laplacian Variance for TQFS to balance efficiency and retrieval effectiveness in practical deployments.

\begin{figure*}[tbp]
    \centering
    \includegraphics[width=0.7\textwidth]{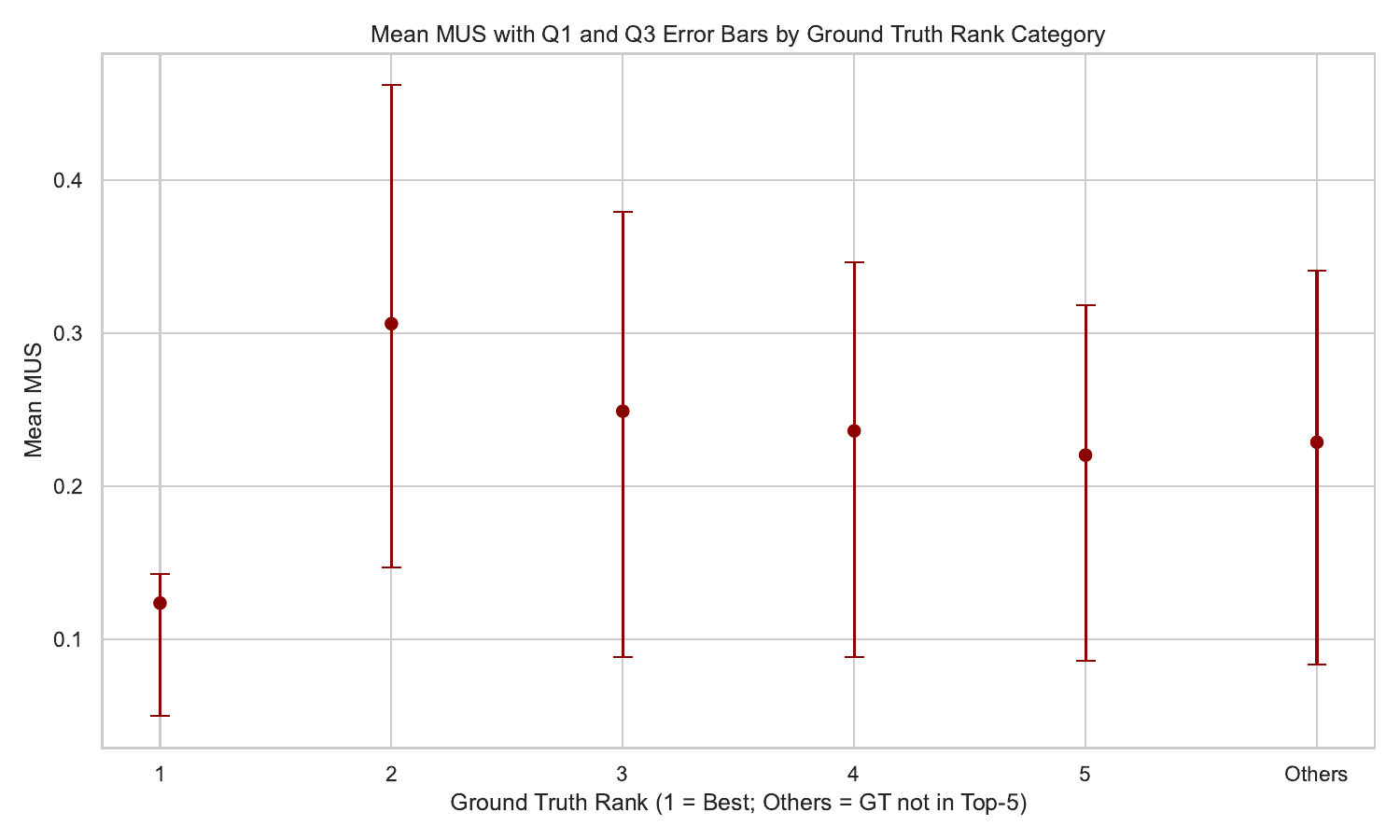}
    \caption{\textbf{Mapping Uncertainty Score (MUS) across different Ground Truth (GT) ranks.} The figure shows the relationship between MUS and the rank position of the ground truth video in the retrieval results. When GT is ranked 1, MUS is consistently low, indicating high confidence in retrieval. However, when GT is ranked 2, MUS is noticeably higher, reflecting the system’s difficulty in distinguishing between the top-ranked candidates. This trend confirms that MUS effectively identifies ambiguous retrieval scenarios, particularly when the correct video is close but not yet ranked first. We set the MUS threshold to 0.2 in UMIVR to trigger interactive refinement in cases of high mapping uncertainty.}
    \label{fig:mus_dis}
\end{figure*} 

\section{Prompt Design for UMIVR}
\label{sec:prompt-design}

We carefully design structured prompts at each stage of the UMIVR framework to systematically address different uncertainty scenarios and improve retrieval accuracy. Specifically, we first introduce detailed yet precise prompts for offline extraction of video meta-information, including video captioning, primary object identification, and semantic scene classification (Table~\ref{tab:prompt_meta_info}). These prompts enforce clear, evidence-based visual descriptions and explicitly discourage speculative content generation.

Subsequently, in the interactive retrieval phase, we propose a hierarchical, uncertainty-guided prompting system to dynamically generate clarifying questions appropriate to query ambiguity levels (Tables~\ref{tab:prompt_lvl0_question},~\ref{tab:prompt_lvl1_question},~\ref{tab:prompt_lvl2_question}). Our prompts progressively transition from open-ended inquiries addressing high textual ambiguity, to targeted visual-distinguishing questions under high mapping uncertainty, and enrichment-oriented queries when uncertainty is minimal. Additionally, simulated user responses are generated through structured prompts that incorporate diverse visual details, enabling realistic iterative refinement and further enhancing retrieval precision (Tables~\ref{tab:answer_prompt},~\ref{tab:query_refinement_prompt}).

In summary, our prompt design provides a principled, flexible, and effective mechanism for managing uncertainty throughout the interactive retrieval process, facilitating clear communication between user and model, and ultimately improving overall retrieval performance.

\clearpage

\section{Limitations}
\label{sec:limitations}

Although our proposed UMIVR framework demonstrates superior performance and strong generalization across multiple retrieval benchmarks, several limitations remain open for future exploration.

First, UMIVR explicitly quantifies uncertainty using external metrics (TAS, MUS, TQFS) rather than relying directly on the inherent uncertainty-awareness capabilities of the underlying Large Language Model (LLM). In other words, current LLMs cannot intrinsically perceive the uncertainty within user queries effectively. If LLMs could internally recognize and quantify these uncertainties, it would enable a more tightly coupled and contextually adaptive generation of clarifying interactions, potentially leading to further improvements in retrieval performance.

Second, our experiments rely on a simulated question-answering mechanism that mimics user responses. While human-simulating question-answering significantly reduces the practical costs of evaluating interactive systems, it inevitably differs from real human interactions in nuanced aspects, such as response variability, hesitation, or misunderstanding. Thus, real-world performance may differ from simulated scenarios, necessitating future validation through human-in-the-loop studies.

Third, the effectiveness of our approach inherently depends on the performance of the underlying multimodal LLM (VideoLLaVA). As current multimodal LLMs still struggle with certain challenging scenarios, such as fine-grained action recognition, temporal understanding, or handling complex textual semantics, improvements in the intrinsic capabilities of VideoLLMs would directly enhance the reliability and overall retrieval performance of our UMIVR framework.

Finally, previous works such as IVR~\cite{ivr_base} have expressed concern regarding potential information leakage when utilizing the same backbone model (e.g., BLIP) for retrieval and VideoQA. However, we found this concern unwarranted in our experiments. Since our retrieval and VideoQA processes are both stateless and executed through independent inference calls, we observed no such leakage effect. Nevertheless, researchers adopting similar strategies should remain cautious and explicitly verify the absence of leakage in different model architectures.

\begin{table*}[htbp]
\centering
\resizebox{1.0\textwidth}{!}
{
\begin{tabular}{l}
\toprule[1pt]
\multicolumn{1}{c}{\textbf{Details about video meta-information generation prompt in UMIVR}} \\ \hline
\begin{tabular}[c]{@{}l@{}}\textbf{System Prompt}:\\ A conversation between a curious human and an AI assistant. The assistant is specialized in analyzing video content and\\ provides detailed, precise, and evidence-based descriptions. Follow these guidelines strictly:\\ - **Precision**: Describe only what is directly observable from the video.\\ - **Detail**: Include all readily visible details while keeping responses focused.\\ - **No Speculation**: If any part of the content is uncertain, explicitly state the uncertainty instead of guessing.\\ \\ \textbf{Caption Prompt}:\\ \{video\_features\}\\ Please provide a detailed and highly accurate caption that fully describes the overall scene or main activity in this video. \\ Make sure your caption includes all relevant visual details and does not exceed 80 words. Do not add any information that \\ is not clearly supported by the video content.\\ \\ \textbf{Main Objects Prompt}:\\ \{video\_features\}Based solely on the visible content of the video, list up to five primary objects or characters you can clearly identify. Each \\ item should be provided as a single word or a brief noun phrase (e.g., 'man', 'tree', 'couch'). Only include items that are \\ explicitly visible and avoid any speculation.\\ \\ \textbf{Scene Type Prompt}:\\ \{video\_features\}\\ Based on the visual content of the video, identify the primary setting, scene type, or dominant visual theme by listing up \\ to five concise keywords (e.g., 'underwater', 'indoor', 'black'). Only include keywords that are directly evident from the \\ video, and do not include any speculative information.\\ \\ \textbf{Max New Tokens}:\\ 1024\\ \\ \textbf{Temperature}:\\ 0.1\end{tabular} \\ \bottomrule[1pt]
\end{tabular}
}
\caption{\textbf{Prompts used in UMIVR for generating video meta-information.} These prompts guide the video LLM in producing accurate and detailed video descriptions, identifying primary objects, and categorizing scene types. The system prompt enforces strict adherence to precision, detail, and avoidance of speculation. The caption prompt ensures comprehensive yet concise descriptions, while the main objects and scene type prompts extract key visual elements.}
\label{tab:prompt_meta_info}
\end{table*}

\begin{table*}[htbp]
\centering
\resizebox{0.6\textwidth}{!}
{
\begin{tabular}{l}
\toprule[1pt]
\multicolumn{1}{c}{\textbf{Details about level-0 question generation prompt in UMIVR}} \\ \hline
\begin{tabular}[c]{@{}l@{}}\textbf{System Prompt}:\\ You are an advanced AI specialized in asking clarifying questions for \\ vague queries. Your task is to extract details—such as appearance, \\ activities, or events—to enable precise retrieval.\\ \\ \textbf{User Prompt}:\\ Query: \{text\_query\}\\ Ask one open-ended clarifying question focusing on the subject's \\ appearance, activities, or events.Return ONLY the question.\\ \\ \textbf{Max New Tokens}:\\ 1024\\ \\ \textbf{Temperature}:\\ 0.1\end{tabular} \\ \bottomrule[1pt]
\end{tabular}
}
\caption{\textbf{Prompt design for Level-0 question generation in UMIVR.} This prompt is used when the initial text query exhibits high text ambiguity, as determined by the Text Ambiguity Score (TAS). The LLM generates an open-ended clarifying question aimed at refining vague queries by eliciting additional details about the subject's appearance, activities, or events. This process helps reduce uncertainty in the retrieval task.}
\label{tab:prompt_lvl0_question}
\end{table*}

\begin{table*}[htbp]
\centering
\resizebox{0.6\textwidth}{!}
{
\begin{tabular}{l}
\toprule[1pt]
\multicolumn{1}{c}{\textbf{Details about level-1 question generation prompt in UMIVR}} \\ \hline
\begin{tabular}[c]{@{}l@{}}\textbf{System Prompt}:\\ You are a clarifying question generator for text-video retrieval. Given \\ a user query and multiple video info, your task is to generate one \\ question that focuses on visual differences. \\ \\ \textbf{User Prompt}:\\ Query: \{text\_query\}\\ Videos: \{video\_meta\_info\_list\}\\ \\ Ask one question starting with What, Where, or Who to distinguish \\ these videos based on visual details.\\ Return ONLY the question.\\ \\ \textbf{Max New Tokens}:\\ 1024\\ \\ \textbf{Temperature}:\\ 0.1\end{tabular} \\ \bottomrule[1pt]
\end{tabular}
}
\caption{\textbf{Prompt design for Level-1 question generation in UMIVR.} This prompt is used when the query has low text ambiguity but exhibits high mapping uncertainty, as determined by the Mapping Uncertainty Score (MUS). Given a user query and multiple retrieved video candidates, the LLM generates a clarifying question that highlights visual distinctions between them. The question is structured to start with "What," "Where," or "Who," ensuring a focus on differentiating key visual elements.}
\label{tab:prompt_lvl1_question}
\end{table*}

\begin{table*}[htbp]
\centering
\resizebox{0.68\textwidth}{!}
{
\begin{tabular}{l}
\toprule[1pt]
\multicolumn{1}{c}{\textbf{Details about level-2 question generation prompt in UMIVR}} \\ \hline
\begin{tabular}[c]{@{}l@{}}\textbf{System Prompt}:\\ You are an advanced AI specialized in asking clarifying questions for queries. Your \\ task is to extract details—such as appearance, activities, or events—to enable \\ precise retrieval.\\ \\ \textbf{User Prompt}:\\ You need to ask a question based on a user query.\\ 1. First you need to evaluate whether the user's query includes sufficient visual details \\ (such as characters, colors, objects, or locations).\\ User Query: \{cur\_text\_query\}\\ \\ 2. Ask a question\\     - If details are missing, generate one question to gather them.\\     - If the query is already detailed, generate a clarifying question to further enrich \\ the description (e.g., 'What other objects are present?', 'What is the main color?', \\ or 'Where is the event taking place?').\\ \\ \\ Return ONLY the question, nothing else.\\ \\ \textbf{Max New Tokens}:\\ 1024\\ \\ \textbf{Temperature}:\\ 0.1\end{tabular} \\ \bottomrule[1pt]
\end{tabular}
}
\caption{\textbf{Prompt design for Level-2 question generation in UMIVR.} This prompt is used when both text ambiguity and mapping uncertainty are low, but further query enrichment is beneficial. The UMIVR evaluates whether the user’s query contains sufficient visual details (e.g., characters, colors, objects, locations). If key details are missing, it generates a question to obtain them; otherwise, it formulates an enrichment-oriented question to enhance the query’s specificity. This iterative refinement helps maximize retrieval accuracy.}
\label{tab:prompt_lvl2_question}
\end{table*}

\begin{table*}[htbp]
\centering
\resizebox{0.68\textwidth}{!}
{
\begin{tabular}{l}
\toprule[1pt]
\multicolumn{1}{c}{\textbf{Details about human-simulation answer generation prompt in UMIVR}} \\ \hline
\begin{tabular}[c]{@{}l@{}}\textbf{System Prompt}:\\ You are a video question answering assistant. When provided with a video and a question, \\ your task is to provide a concise, one-sentence answer. Your answer should clearly state \\ the key visual details such as people, objects, scenes, and events. Keep it clear, direct, and \\ focused on essential information.\\ \\ \textbf{User Prompt}:\\ \{video\_features\}\\ \\ Question: \{question\}\\ \\ Provide a one-sentence answer that clearly identifies the key visual details in the video, \\ such as people, objects, scenes, and events.\\ \\ \textbf{Max New Tokens}:\\ 1024\\ \\ \textbf{Temperature}:\\ 0.7\end{tabular} \\ \bottomrule[1pt]
\end{tabular}
}
\caption{\textbf{Prompt design for human-simulation answer generation in UMIVR.} This prompt is used to simulate user responses in interactive retrieval by generating concise, one-sentence answers to clarifying questions based on video content. The video LLM extracts key visual details, including people, objects, scenes, and events, ensuring clarity and relevance. A higher temperature setting (0.7) is used to introduce variability, better mimicking the natural diversity in human responses. This simulation enables iterative query refinement, improving retrieval accuracy through realistic user interactions.}
\label{tab:answer_prompt}
\end{table*}

\begin{table*}[htbp]
\centering
\resizebox{0.65\textwidth}{!}
{
\begin{tabular}{l}
\toprule[1pt]
\multicolumn{1}{c}{\textbf{Details about query refinement prompt in UMIVR}} \\ \hline
\begin{tabular}[c]{@{}l@{}}\textbf{System Prompt}:\\ You are an expert in query refinement for interactive text-video retrieval. \\ Your task is to synthesize and update a previous query with new details \\ from the current answer. Ensure the new query includes key information \\ (e.g., characters, events, objects, colors, locations) and does not exceed \\ 60 words.)\\ \\ \textbf{User Prompt}:\\ Previous Query: \{pre\_query\}\\ \\ Current Answer (includes new information to enhance video retrieval):\\ \{cur\_answer\}\\ \\ Combine the above into one concise, positive declarative sentence that \\ includes key details (characters, events, objects, colors, locations, etc.). \\ Ensure the new query leverages the new information from the current \\ answer for better retrieval and is no longer than 60 words.\\ \\ Only return the refined query, nothing else.\\ \\ \textbf{Max New Tokens}:\\ 1024\\ \\ \textbf{Temperature}:\\ 0.1\end{tabular} \\ \bottomrule[1pt]
\end{tabular}
}
\caption{\textbf{Prompt design for query refinement in UMIVR.} This prompt is used to iteratively improve user queries by incorporating new details extracted from simulated user answers. The LLM synthesizes the previous query with newly provided information (e.g., characters, events, objects, colors, locations), ensuring a more precise and informative query for video retrieval.}
\label{tab:query_refinement_prompt}
\end{table*}


\end{document}